\documentclass[nonblindrev]{informs3noheader}

\OneAndAHalfSpacedXI

\usepackage{graphicx}
\usepackage{multirow}
\usepackage{url}

\usepackage[breaklinks,hypertexnames=false]{hyperref}
\usepackage[dvipsnames]{xcolor}
\usepackage{booktabs}
\usepackage{tabularx}
\usepackage{caption}
\usepackage[size=footnotesize]{subcaption}
\usepackage[authoryear,sort]{natbib}
\bibpunct[, ]{(}{)}{,}{a}{}{,}%
\def\bibfont{\small}%

\definecolor{myblue}{RGB}{0,0,150}
\hypersetup{
	colorlinks=true,
	linkcolor=MidnightBlue,
	filecolor=blue,
	citecolor=myblue,
	urlcolor=myblue
}

\newcommand{\ignore}[1]{}

\newcommand{\haiku}{\texttt{claude-haiku-4-5}}
\newcommand{\sonnet}{\texttt{claude-sonnet-4-6}}
\newcommand{\nano}{\texttt{gpt-4.1-nano}}
\newcommand{\mini}{\texttt{gpt-5.4-mini}}
\newcommand{\sol}{\texttt{gpt-5.6-sol}}
\newcommand{\llamafast}{\texttt{llama-3.1-8b}}
\newcommand{\llama}{\texttt{llama-3.3-70b}}
\newcommand{\llamafour}{\texttt{llama-4-scout}}
\ABSTRACT{

 Large language models are increasingly used to support organizational decisions. Despite their rapid and widespread adoption, users often lack a principled basis for assessing whether to rely on a specific recommendation. We introduce \emph{epistemic warrant}, a decision-level construct that characterizes the scope over which a recommendation is supported. We operationalize this construct through a reliance certificate that uses behavioral evidence from four tiers of tests to assess a recommendation's underlying warrant and assign it to a warrant category. We assess this operationalization using complementary forms of construct-validity evidence, including transformation validity, known-groups tests, independent human consensus, discriminant analyses, and robustness across models and alternative implementations. The results show that the certificate behaves in theoretically expected ways: its transformations instantiate their intended relationships, certificate assignments increase across groups prespecified to represent increasing levels of warrant, and stronger assigned warrant is associated with greater independent human consensus. Certificate assignments also capture information beyond verbalized confidence and are not readily explained by decision difficulty. The main conclusions remain robust across models and operationalizations. These findings support the reliance certificate as an operationalization of epistemic warrant for assessing the scope of support for a particular LLM recommendation when ground truth is unavailable.
 }

\TheoremsNumberedThrough
\EquationsNumberedThrough
\begin{document}
	
	\TITLE{Epistemic Warrant for LLM Recommendations: Characterizing the Basis for Reliance When Ground Truth Is Unavailable}
	\RUNTITLE{Epistemic Warrant}
	\RUNAUTHOR{Vardi and Sedoc}
	\ARTICLEAUTHORS{%
		\AUTHOR{Shai Vardi}
		\AFF{University of South Florida, Muma College of Business, \EMAIL{vardi@usf.edu}}
		\AUTHOR{Jo\~ao Sedoc}
		\AFF{New York University, \EMAIL{jsedoc@nyu.edu}}
	}
	\maketitle
	\pagestyle{plain}

\section{Introduction}

When is a decision-maker justified in acting on a recommendation from a
large language model (LLM)? This question has practical urgency as
organizations are increasingly integrating LLMs
into their business workflows~\citep{arslan2026large}.

LLMs are being adopted despite growing evidence that LLM recommendations can be unreliable. LLM judgments can be sensitive to seemingly irrelevant contextual features~\citep{valkanova2024irrelevant,yang2024llm,ceballos2024open}, exhibit systematic biases related to demographic attributes~\citep{an2025measuring,wen2025faireassessingracialgender,templin2025framework,hofmann2024ai}, and contain incorrect responses~\citep{huang2025survey}. These concerns are even more consequential as LLMs are integrated into
high-stakes domains such as medical triage and hiring
\citep{ramaswamy2026chatgpt,wang2024jobfair}. 
A confounding difficulty is that in many scenarios when LLMs are applied, there is no ground truth against which a recommendation can be verified.

Prior work has proposed various methods for assessing and improving LLM outputs. Confidence methods estimate the model's certainty about an output, and calibration methods assess whether that certainty corresponds to correctness~\citep{guo2017calibration,kadavath2022language,tian2023just}. Repeated sampling is used to evaluate the model's consistency across generations~\citep{manakul2023selfcheckgpt}, and selecting the most frequent response can improve accuracy~\citep{wang2023selfconsistency}. Metamorphic testing examines whether responses satisfy expected relations across systematically transformed prompts~\citep{chen2020metamorphic,cho2025metamorphic}. Recent work addresses the problem that prompts are often underspecified by identifying missing information, eliciting clarification, or evaluating responses once additional context is supplied~\citep{zhang2024clamber,malaviya2025contextualized}. 
A complementary IS literature examines why users adopt systems, trust
advice, follow recommendations, or abandon algorithms after observing
them err
\citep{davis1989perceived,venkatesh2003user,
dietvorst2015aversion,logg2019appreciation}. These literatures address whether a model's output is likely correct, repeatable, or robust; how it can be improved; and when people choose to rely on it. However, they do not characterize how broadly and under what conditions reliance on a particular recommendation is supported.

To address this gap, we introduce \emph{epistemic warrant}, a decision-level construct for characterizing the scope of support for relying on a specific recommendation. In epistemology, warrant is treated as a graded property of a belief: when a belief is true and possesses sufficient warrant, it qualifies as knowledge~\citep{plantinga1993warrant}. We adapt this idea from beliefs to recommendations. Warrant is defined with respect to a recommendation, its focal context, and reasonable refinements and extensions of that context.\footnote{We intentionally do not attempt to define ``reasonable''. This follows common practice in epistemology, where related notions such as nearness and ease are likewise left undefined~\citep{williamson2000knowledge,sosa1999defeat}; see Section~\ref{sec:theoretical_foundations}.} A refinement makes the context more specific; an extension broadens or shifts it to a related context. We  distinguish between four categories of warrant for a recommendation with respect to a focal context. 
1) \emph{No Warrant}: the recommendation is not supported in the focal context; 2) \emph{Conditional Warrant}: the recommendation is supported in the focal context but not under all  reasonable refinements of the context; 3) \emph{Basic Warrant}: the recommendation is supported over the focal context and all reasonable refinements, but not over some reasonable extensions; 4)  \emph{Strong Warrant}:  the recommendation is supported over the focal context, all reasonable refinements, and all reasonable extensions. 
The four categories capture progressively broader scope of support for a recommendation.

Epistemic warrant is a theoretical property of a recommendation; it cannot be directly observed. We therefore develop a \emph{reliance certificate}: a behavioral operationalization of epistemic warrant. The reliance certificate is derived from observable model behavior and consists of four tiers of tests, denoted T0--T3. T0 tests stochastic stability by asking whether the model reproduces its preference across repeated generations of the original prompt. T1 tests invariance under transformations that preserve the underlying decision, such as reordering the alternatives or making semantically equivalent substitutions. T2 tests whether the preference holds across refinements of the stated decision. T3 tests whether the preference extends beyond the focal context to broader or substantively related contexts. A change in the model's implied preference constitutes a violation at the corresponding tier.
The reliance certificate maps the outcomes of these tests to the four warrant categories. A violation at T0 or T1 results in a \emph{No Warrant} assignment. A recommendation that passes T0 and T1 but violates T2 is assigned \emph{Conditional Warrant}. A recommendation that also passes T2 but violates T3 is assigned \emph{Basic Warrant}, while one that passes all four tiers is assigned \emph{Strong Warrant}.

The certificate assignment depends on both the specific model used and the particular implementation of the tests. The same recommendation may therefore be assigned different categories across models or implementations. We do not assume that a certificate assignment perfectly corresponds to the underlying warrant. Moreover, because warrant is unobservable, the correspondence between certificate assignments and warrant cannot be verified directly. We therefore evaluate the construct validity of the certificate: whether empirical evidence and theoretical rationales support interpreting certificate outcomes as informative about epistemic warrant \citep{messick1989meaning,messick1995validity}. 

To evaluate construct validity empirically, we need a concrete implementation of the reliance certificate. The certificate admits multiple implementations consistent with its specification. Our aim is to assess whether the resulting assignments exhibit the patterns theoretically expected of epistemic warrant; if the certificate captures the intended construct, the validity evidence should persist across reasonable implementations.  We therefore select one implementation for our validation and use a second implementation as a robustness check.

We implement the certificate-assignment procedure for pairwise recommendations, in which the model chooses between two alternatives. Pairwise choice allows us to examine model behavior under controlled changes to the decision problem. In addition, a growing body of work finds pairwise evaluations to be more consistent, robust, and aligned with human judgments than absolute scores~\citep{liu2024aligning,qin2023large,zheng2023judging}.
We generate the certificate using an automated pipeline. Given an original pairwise prompt, an auxiliary LLM generates tier-specific T1--T3 transformations. The focal model is then queried repeatedly on the original prompt for T0 and on each generated transformation for T1--T3. The resulting test outcomes are mapped to a warrant category using the hierarchical classification rule. 

Our validation strategy follows established guidance for construct validation in IS research \citep{MacKenzie2011Construct}. We assess content validity by examining whether the generated transformations instantiate their intended tier relationships. Our known-groups validity tests check whether certificate assignments distinguish recommendations prespecified by domain experts to differ in warrant. We assess nomological validity by examining whether certificate assignments exhibit the theoretically expected relationship with an external variable, for which we use independent human consensus. Human consensus is theoretically related to warrant because recommendations supported across a broader range of reasonable contextual interpretations should elicit greater agreement among independent decision makers. We examine discriminant validity by assessing whether the assignments capture information distinct from related measures, specifically verbalized confidence and decision difficulty. Finally, we assess generalizability and robustness by examining whether the validity evidence persists across models and alternative implementations of the certificate. Table~\ref{tab:validation_framework} summarizes the question addressed by each component of this strategy and the evidence used to evaluate it.

\newcommand{\skipamount}{1pt}
{\OneAndAHalfSpacedXI
\begin{table}
\centering
\captionof{table}{Validation Framework for the Reliance Certificate}
\label{tab:validation_framework}

{\footnotesize
\renewcommand{\arraystretch}{1.2}
\setlength{\tabcolsep}{4pt}

\begin{tabular}{
>{\raggedright\arraybackslash}p{0.17\textwidth}
>{\raggedright\arraybackslash}p{0.39\textwidth}
>{\raggedright\arraybackslash}p{0.39\textwidth}
}
\noalign{\vskip \skipamount}
\hline
\noalign{\vskip \skipamount}
\textbf{Component}
&
\textbf{Assessment Question}
&
\textbf{ Evidence}
\\
\noalign{\vskip \skipamount}
\hline
\noalign{\vskip \skipamount}

Content validity
&
Do the generated transformations faithfully instantiate their intended tier relationships?
&
Independent human classification of sampled prompt transformations
(Section~\ref{sec:indicator_validation_results}).
\\
\noalign{\vskip \skipamount}
\hline
\noalign{\vskip \skipamount}

Known-groups validity
&
Does the certificate distinguish among groups prespecified to represent increasing levels of warrant?
&
Ordered-trend and correspondence tests against prespecified warrant categories
(Section~\ref{sec:known_groups_results}).
\\
\noalign{\vskip \skipamount}
\hline
\noalign{\vskip \skipamount}

Nomological validity
&
Do certificate assignments exhibit the theoretically expected relationship with an external variable?
&
Association between certificate-assigned warrant and independent human consensus
(Section~\ref{sec:nomological_results}).
\\
\noalign{\vskip \skipamount}
\hline
\noalign{\vskip \skipamount}

Discriminant validity
&
Do certificate assignments capture information distinct from related measures?
&
Divergence from verbalized confidence, explanatory value for human consensus beyond confidence, and distinction from decision difficulty
(Section~\ref{sec:discriminant_results}).
\\
\noalign{\vskip \skipamount}
\hline
\noalign{\vskip \skipamount}

Generalizability and robustness
&
Do the substantive conclusions persist across models and alternative implementations of the certificate?
&
Cross-model comparisons and robustness analyses
(Section~\ref{sec:robustness_results}).
\\
\noalign{\vskip \skipamount}
\hline
\end{tabular}
}
\end{table}}

Our empirical evaluation uses two datasets: 100 pairwise recommendation prompts spanning 22 topics and a known-groups dataset of 40 prompts whose warrant categories are assigned ex ante by subject-matter experts or independently validated by domain experts. We use the first dataset to assess content, nomological, and discriminant validity, as well as generalizability and robustness. We use the second to assess known-groups validity. We evaluate LLMs from the OpenAI, Anthropic, and Llama families.

To evaluate content validity, we use human raters to assess whether sampled transformations instantiate their intended tier relationships, and find that independent raters' majority classifications match the intended tier for 23 of 24 sampled transformations. In the known-groups test, the mean certificate-assignment was non-decreasing across the prespecified warrant categories for all six models tested, and Page's test supported the expected ordered trend for five of the six. This provides evidence of known-groups validity: on average, the certificate assigns higher warrant categories to recommendations independently prespecified to have higher warrant. For nomological validity, we found that stronger certificate-assigned warrant was positively associated with independent human consensus for all seven models and significantly for six, and this association persists when controlling for decision difficulty. Our discriminant validity analysis provides evidence that warrant captures information distinct from verbalized confidence.  Our conclusions remain robust across models and alternative operationalizations.

This paper makes three contributions. First, it develops epistemic warrant as a decision-level construct that characterizes the scope of support for relying on a particular LLM recommendation.  The construct is distinct from related constructs such as correctness, confidence and trust. It distinguishes recommendations that lack support in the focal context, are supported only under some refinements of the focal context, are supported throughout this context, or remain supported beyond it. Second, it develops a reliance certificate that operationalizes the different warrant categories through four tiers of behavioral tests. Third, it provides complementary forms of construct-validity evidence supporting the interpretation of certificate assignments as informative about epistemic warrant. The findings are robust and persist across models and alternative implementations. 

\section{Conceptual Foundations and Related Literature}\label{sec:related_lit}

Our work connects three streams of literature: epistemic theories of warrant and justification, IS research on trust and reliance in decision support, and research on robustness, consistency, and behavioral evaluation of LLMs.

\subsection{Theoretical Foundations}
\label{sec:theoretical_foundations}

Our conception of epistemic warrant retains two features of \citet{plantinga1993warrant}'s account: warrant comes in degrees, and it is distinct from truth. It differs in two respects. First, Plantinga's warrant is a property of a belief, whereas ours is a property of a recommendation relative to its decision context. Second, in Plantinga's account, sufficient warrant, together with truth, yields the binary status of knowledge: a person either knows \(p\) or does not. In our setting, objective correctness may be unavailable or undefined, and degrees of warrant are themselves the outcome of interest, each corresponding to a different scope of support for the recommendation.

Our adaptation is motivated primarily by counterfactual approaches in epistemology. A central idea in these approaches is that stronger epistemic standing derives from support that persists across a broader range of relevant alternative circumstances. \citet{nozick1981philosophical} asks whether an agent would cease to believe a proposition in nearby circumstances in which it is false. \citet{sosa1999defeat} and \citet{williamson2000knowledge} instead require that a belief be \emph{safe}: it could not easily have been false, so that it remains true across nearby circumstances. \citet{derose1995solving} makes such support graded, arguing that the further from the actual situation one's belief continues to match the truth, the stronger one's epistemic position. We adopt this idea directly: a recommendation's warrant increases with the range of reasonable variations of its decision context across which it remains supported.

The warrant categories are distinguished by whether a recommendation remains supported across reasonable variations of its decision context. We treat \emph{support} as a primitive relation rather than attempting to reduce it to more basic concepts, and leave \emph{reasonable} open-textured. This is consistent with approaches in philosophy that build theories around concepts that are not formally defined.  \citet{williamson2000knowledge} treats knowledge as a primitive and relies on the notion of close possibilities without providing a  formal definition  of closeness; \citet{nozick1981philosophical} relies on an intuitive understanding of subjunctive conditionals, and does not define nearness;  
and \citet{lewis2013counterfactuals} declines to reduce similarity between possible situations to a precise metric. 
 In each case, the theoretical structure remains informative without requiring a unique formalization of the underlying notion. Likewise, the structure of epistemic warrant does not depend on a particular definition of support or a particular formalization of reasonableness. Importantly though, whether a variation is reasonable must be determined independently of whether the recommendation survives it.

\subsection{Trust and Reliance in Decision Support}

There is a long line of research on trust in decision-support systems. Early work on decision aids shows that such systems do not simply provide information, but alter the cognitive effort associated with different decision strategies and thereby shape how users search for, process, and use information \citep{todd1991decisionaids,todd1992information}. Subsequent work identifies trust as a central mechanism shaping adoption and reliance on recommendation agents \citep{benbasat2005trust,komiak2006personalization}, and shows that explanations and attributional cues can strengthen users' trust in system recommendations \citep{wang2007explanation,wang2008attributions}. More recently, the rise of opaque AI systems has pushed this literature toward organizational questions: how firms govern algorithmic decision-making, and how managers maintain meaningful oversight when the underlying logic is inaccessible \citep{berente2021managing}.

A central theme across this literature is that reliance must be calibrated. Users may rely inappropriately when trust exceeds warranted system capability \citep{lee2004trust}, under-rely on algorithms after observing errors \citep{dietvorst2015aversion}, or over-rely on algorithmic advice relative to human judgment \citep{logg2019appreciation}. However, this literature primarily explains how users form trust in algorithmic systems and how trust affects reliance,  not on how to evaluate whether a specific recommendation is itself worthy of reliance.

\subsection{Behavioral Evaluation of LLM Outputs Without Ground Truth}

When objective ground truth is unavailable or difficult to obtain, prior research has used patterns in a model's behavior across repeated or systematically varied inputs as  signals about its outputs.

One approach is to query the model multiple times and examine the consistency of its responses. Self-consistency methods generate multiple responses to the same problem and select the answer that appears most often across them \citep{wang2023selfconsistency}. Consistency across repeated samples can also be used to identify potentially non-factual claims, as in SelfCheckGPT \citep{manakul2023selfcheckgpt}. These approaches show that repeated agreement contains useful information, but they generally use it to improve an answer or identify unreliable content.

Metamorphic testing evaluates outputs by testing expected relations across systematically related inputs \citep{chen2020metamorphic}. Recent applications show that LLMs frequently violate such relations across transformed prompts \citep{cho2025metamorphic,de2026semantic,li2024drowzee}. Related studies find that model judgments can change in response to features that should not alter the underlying decision, including option order, anchoring, decoy alternatives, and verbosity \citep{zheng2023judging,yin2025fragile,lou2024anchoring,valkanova2024irrelevant,saito2023verbosity}. These forms of fragility cannot always be identified from the model's own explanations, which may not faithfully reflect the factors driving its answer \citep{turpin2023language,mayne2025llms}.

Another stream addresses underspecified prompts by identifying missing information, requesting clarification, or evaluating responses after additional context is supplied \citep{zhang2024clamber,malaviya2025contextualized}. These approaches primarily seek to improve the response or assess whether the model uses additional information appropriately. Our concern is different: we vary the context to determine the scope over which the original recommendation remains supported. Our framework distinguishes changes under refinements of the focal context from changes under extensions beyond it, using this distinction to characterize conditional support and the boundaries of reliance.

A different signal is the model's expressed confidence in its answer. Prior work shows that, on tasks where correctness can be independently assessed, verbalized confidence can predict factual correctness and may be better calibrated than confidence inferred from token probabilities \citep{kadavath2022language,tian2023just}. Such validation necessarily depends on an independent criterion of correctness, which may be unavailable for the kinds of recommendations considered here. Moreover, models are often overconfident, and the quality of their confidence estimates varies across models, tasks, and elicitation procedures \citep{xiong2024can,zhao2024fact}. Confidence therefore captures the model's stated certainty, whereas our concern is whether and over what scope the recommendation is behaviorally supported. A model may express high confidence yet reverse its recommendation under repeated sampling, decision-preserving reformulations, or plausible contextual refinements. Confidence alone therefore cannot establish either the strength or the contextual scope of epistemic warrant.

Our framework draws on several of these approaches. T0 is related to repeated-sampling and self-consistency methods, while T1 is related to decision-preserving transformations used in metamorphic testing. T2 and T3 use contextual variation to characterize the scope of support for an already-issued recommendation, although we extend the use of context beyond identifying missing information or eliciting clarification. We integrate these signals into a decision-level framework in which different patterns of stability and change have different implications for warrant. The reliance certificate operationalizes this framework by mapping observable model behavior to assignments intended to characterize the scope of support for a particular recommendation.

\section{Theoretical Development of Epistemic Warrant}
\label{sec:framework} 
In this section, we define epistemic warrant, specify the four warrant categories, and describe how the reliance certificate uses a model's observable behavior to estimate them.

\subsection{Epistemic Warrant and Scope of Support}
\label{sec:object}

Consider a pairwise decision problem \(d=(a,b,S)\), where \(a\) and \(b\) are the alternatives and \(S\) is the focal decision context.  
 A \emph{refinement} of $S$ is a narrower context \(S'\subsetneq S\) representing a more specific case contained within the focal context. An \emph{extension} is a context beyond the focal one: either a broader context \(S'\) with \(S\subsetneq S'\), or a substantively related context for which neither contains the other. Let \(\mathcal{R}(S)\) and \(\mathcal{E}(S)\) denote the sets of reasonable refinements and reasonable extensions of \(S\), respectively.

Let \(r\in\{a,b\}\) be a  recommendation for \(d=(a,b,S)\).  We treat whether a recommendation is \emph{supported} in a given context as a primitive relation; epistemic warrant characterizes the scope over which this relation holds.  We define four warrant categories for $r$:

\begin{enumerate}
\item \textbf{No Warrant:} \(r\) is not supported in the focal context \(S\).
\item \textbf{Conditional Warrant:} \(r\) is supported in \(S\), but not in some reasonable refinement \(S'\in\mathcal{R}(S)\).
\item \textbf{Basic Warrant:} \(r\) is supported in \(S\) and in every reasonable refinement, but not in some reasonable extension \(S'\in\mathcal{E}(S)\).
\item \textbf{Strong Warrant:} \(r\) is supported in \(S\), in every reasonable refinement, and in every reasonable extension.
\end{enumerate}

A recommendation supported in the focal context may not be supported in all of its reasonable refinements. Such a reversal does not imply that the recommendation is unsupported for the original context; rather, it identifies a condition under which its support changes. 
 A reversal in a reasonable extension does not weaken the support for the focal decision; however, it identifies a boundary beyond which the recommendation should not be generalized. 


\subsection{Operationalizing Epistemic Warrant}
\label{subsec:four_tiers}

 Let \(Q(d)\) denote the set of natural-language statements that express decision problem \(d\), let \(m\) be an LLM, let \(q\in Q(d)\) be the focal prompt presented to \(m\), and let \(r_m(q)\) denote the model's recommendation. The epistemic warrant of $r_m(q)$ cannot be observed directly. We therefore operationalize
it using behavioral evidence from the model. 

The operationalization consists of  four tiers of tests.  
T0 and T1 provide evidence about whether the recommendation is supported in the focal context by testing whether the model's preference is stable across repeated generations and invariant to decision-preserving changes in formulation. T2 and T3 provide evidence about the scope of that support by testing whether it persists across reasonable refinements and reasonable extensions of the focal context, respectively. 

\paragraph{Tier 0: Stochastic stability.}
Tier~0 repeatedly queries the model with the focal prompt under stochastic generation. A T0 violation occurs when the model fails to reproduce its preference.

\paragraph{Tier 1: Decision-preserving invariance.}
Tier~1 evaluates alternative prompts \(q'\in Q(d)\). These probes express the same decision problem while changing how it is presented, for example through paraphrasing or reversing the order of the alternatives. A T1 violation occurs when \(r_m(q')\neq r_m(q)\) for a valid decision-preserving transformation \(q'\).

\paragraph{Tier 2: Scope refinement.}
Tier~2 evaluates decision problems \(d'=(a,b,S')\) for a sample of refinements \(S'\in\mathcal{R}(S)\). A T2 violation occurs when the model recommends a different alternative in one of these narrower contexts.

\paragraph{Tier 3: Scope extension.}
Tier~3 evaluates decision problems \(d'=(a,b,S')\) for a sample of extensions \(S'\in\mathcal{E}(S)\), including both broader and substantively related contexts. A T3 violation occurs when the model recommends a different alternative in one of these contexts.

{\OneAndAHalfSpacedXI
\begin{figure}[t]
\centering
\includegraphics[width=0.42\textwidth]{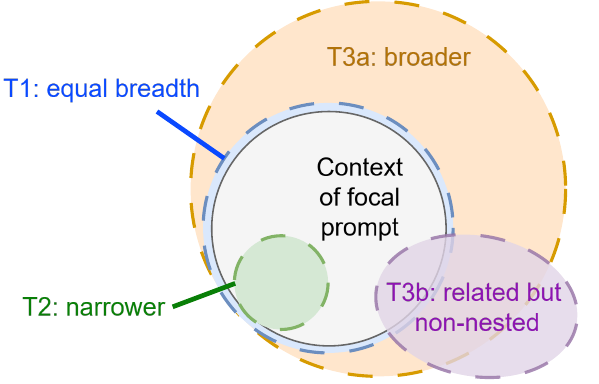}
\caption{Conceptual relationships among the contextual transformation tiers. Tier~1 transformations preserve  the same contextual breadth as the focal decision; Tier~2 transformations evaluate narrower subcontexts; and Tier~3 transformations evaluate either broader contexts or substantively related contexts that are not nested within the focal context. Tier~0 is not shown because it consists of repeated queries of the unchanged focal prompt rather than a contextual transformation.}
\label{fig:context_tiers}
\end{figure}}

Figure~\ref{fig:context_tiers} illustrates these relationships. Suppose the focal context concerns \emph{managers}. Alternative formulations concerning managers preserve the focal context. \emph{HR managers} constitutes a refinement. \emph{Employees} provide a broader extension, while \emph{HR employees} provide a substantively related extension that overlaps with, but is not nested within, the focal context.

\subsection{The Reliance Certificate and Category Assignment}
\label{subsec:categories}

Let \(n_k\) denote the number of valid probes evaluated at tier \(k\), and let \(v_k\) denote the number of observed preference reversals. The certificate for recommendation \(r_m(q)\) records
\[
C_m(q)=\bigl((n_0,v_0),(n_1,v_1),(n_2,v_2),(n_3,v_3)\bigr).
\]

We use a hierarchical, noncompensatory classification rule to map these observed outcomes to the four theoretical warrant categories. The hierarchy follows the conditions defining the warrant categories: evidence of support in refinements or extensions cannot compensate for a lack of support in the focal context, and support in extensions does not compensate for a lack of support in a reasonable refinement. Tier $k$ is passed when \(v_k=0\). The recommendation \(r_m(q)\) is assigned \emph{No Warrant} if there is a T0 or T1 violation; \emph{Conditional Warrant} if it passes T0 and T1 but violates T2; \emph{Basic Warrant} if it passes T0--T2 but violates T3; and \emph{Strong Warrant} if it passes all four tiers with no violations. 

The certificate assignment and the underlying theoretical category are distinct. The theoretical categories are defined over the full set of reasonable refinements and extensions, whereas the certificate can examine only a finite set of probes. The certificate is also model-specific, whereas the theoretical warrant of a recommendation is not defined in terms of the model used to  evaluate it. A Strong Warrant assignment, for example, establishes that no violation was observed among the probes examined; it does not demonstrate that the recommendation would remain supported under every reasonable variation. More generally, model behavior provides evidence about support rather than directly revealing it. The certificate therefore provides empirical evidence about a recommendation's underlying warrant category rather than a direct observation of that category.

Although we operationalize each tier as pass/fail, the certificate retains both \(n_k\) and \(v_k\).  Therefore, the categorical assignment can  be supplemented with tier-specific violation rates and the particular contexts in which reversals occur. We discuss the implications of using this additional granularity in Section~\ref{sec:discussion}.


\section{Validation Design}


In this section, we first describe the certificate-generation and querying procedure used to implement the reliance certificate. We then present the empirical validation designs and analyses used to assess content validity, known-groups validity, nomological validity, discriminant validity, and generalizability and robustness.

\subsection{Certificate Implementation}\label{design:certificate_implementation}

We implemented the reliance certificate as a predetermined querying pipeline that can be applied to any binary recommendation prompt. Given a focal prompt, \haiku~generates the associated Tier~1--3 queries using tier-specific transformation instructions, while Tier~0 is assessed by repeatedly querying the unchanged focal prompt. Haiku was selected because it provided reliable instruction following at relatively low cost; we performed a robustness analysis using an alternative model, described in Section~\ref{prompt_evaluation}.

To construct a certificate, a target model is first queried on the focal prompt to obtain a reference recommendation and is then queried on the associated Tier~0--3 certificate queries. 
All violations are recorded, and the resulting tier-level violation profile is mapped to the four warrant categories using the hierarchical rules defined in Section~\ref{subsec:categories}. Appendix~\ref{app:certificate_implementation} provides the transformation instructions, querying specifications, filtering and answer-mapping rules, and the complete procedure for constructing the tier-level violation profile. The corresponding implementation code is available in the replication repository (Section~\ref{sec:data_code}).

\subsection{Empirical Validation Design}

\subsubsection{Prompt Development and Sampling.} \label{prompt_evaluation}

We constructed 100 binary recommendation prompts from 22 base topics drawn from LMSYS Chatbot Arena~\citep{zheng2024lmsys} and WildChat~\citep{zhao2024wildchat}. Base topics were selected according to four criteria: they were accessible to a general audience, concerned everyday or professional decisions, could be expressed as a choice between two alternatives, and lacked a readily verifiable ground-truth answer. For each base topic, we constructed four or five related prompts intended to span different expected levels of epistemic warrant.

The prompt set, transformation procedure, and warrant rules were fixed before evaluating the focal models. For each focal prompt, we generated eight Tier~2 and eight Tier~3 transformations using the predetermined Haiku-based pipeline and used all generated transformations without post-generation screening. Tier~1 consisted of option permutation, valence inversion,\footnote{Valence inversion reverses the evaluative direction of the prompt while leaving the alternatives and substantive context unchanged. For example, a question asking which option is \emph{better} may be reformulated to ask which is \emph{worse}, so that the corresponding preferred alternative should reverse if the underlying preference is preserved.} their combination, and semantically close contextual substitutions. Contextual substitutions were retained as Tier~1 transformations only when their GloVe cosine distance was no greater than 35 on the 0--200 distance scale, yielding an average of 4.2 Tier~1 transformations per prompt. The complete transformation procedure is provided in Appendix~\ref{app:certificate_implementation}.

The same fixed set of 100 focal prompts was used in the content-validity, nomological-validity, and discriminant-validity analyses. The focal models and human participants were evaluated independently; human participants were not shown model recommendations or certificate classifications.


In addition to the primary Haiku-based transformation procedure, we developed  an alternative transformation-generation procedure for robustness analyses. The alternative procedure uses \mini, revised
tier-specific instructions, and exactly eight transformations per tier. We  applied this procedure to the same 100 focal prompts for \haiku, \sonnet, \nano, \mini, \sol\ and \llamafour. Appendix~\ref{app:alternative_procedure} details the
differences between the two transformation procedures.

\subsubsection{Content-Validity Design.}
\label{design:indicator}

To assess whether the generated transformations faithfully instantiated their intended tiers, we recruited 21 participants from Prolific to compare focal recommendation prompts with transformed versions. We used stratified random sampling, selecting eight prompt pairs from each of Tiers~1--3, for a total of 24 pairs. Three participants were excluded for failing an attention check, leaving 18 participants for analysis.\footnote{We applied the same participant-selection criteria in the content-validity and human-consensus experiments.} Each retained participant evaluated all 24 pairs, yielding 144 ratings per tier.

Participants classified the relationship between each focal and transformed prompt as approximately equal in breadth, narrower, broader, related but non-nested, or indeterminate.\footnote{We used ``approximately equal breadth'' instead of ``the same decision'' as the latter is ambiguous, particularly for valence-inverted questions, where the underlying preference is preserved but the answer should reverse.}  Approximately equal breadth corresponds to Tier~1, narrower corresponds to Tier~2, and broader or related but non-nested correspond to Tier~3. Complete instructions and response options are provided in Appendix~\ref{app:indicator_prompts}. Participants were not shown model responses, certificate outcomes, or the intended tier of the transformation.

\subsubsection{Model Evaluation.}

We evaluated the certificate on seven language models spanning three model families: \haiku, \sonnet, \nano, \mini, \sol, \llamafast, and \llama. Each model was evaluated on the same fixed set of 100 focal prompts and their associated certificate queries.

For each focal prompt, we first obtained a reference recommendation by instructing the model to return exactly one of the two response options  at $\tau=0$. Tier~0 was then evaluated using 10 additional queries of the unchanged focal prompt at $\tau=1$. Tiers~1--3 were evaluated by querying each associated transformation once at $\tau=0$. The exception is \sol, for which no temperature parameter was supplied because the model does not support manual temperature configuration under the evaluated setting. If a model failed to return one of the two permitted response options, it was reminded of the required format and queried again, up to four additional attempts.

A violation was recorded when the recommendation on a certificate query differed from the reference recommendation, after applying the appropriate answer mapping for valence-inverted Tier~1 transformations. The resulting tier-level violation profile was mapped to the four warrant categories using the predetermined hierarchical classification rule.

For each focal prompt, we additionally elicited verbalized confidence
immediately after obtaining the reference recommendation, within the same
conversation. Models were asked, ``On a scale from 0 to 100, how confident are
you in that answer? Respond with only the number.'' Responses were required to
be whole numbers between 0 and 100; nonconforming responses were re-queried
using the same retry procedure. 

\subsubsection{Human Consensus Experiment.}
\label{design:nomological}

To assess nomological validity, we use independent human consensus as a theoretically related but conceptually distinct external variable. 
The intuition for the expected correlation is as follows. Under No Warrant, the recommendation lacks support in the focal context; there is therefore the least reason to expect human decision makers to agree. Under Conditional Warrant, the recommendation is supported in the focal context, which should produce greater consensus; however, decision makers may implicitly focus on different refinements or extensions, for example by drawing on personal experience. Consider the question, “Is a laptop or a tablet preferable for college students?” A decision maker who knows an engineering student might implicitly focus on engineering students, a refinement of the focal context. Another who does not know any college students but has a child in high school might draw on that experience instead. High-school students constitute a related, non-nested context and therefore fall under T3. The source of disagreement arising from different refinements is removed under Basic Warrant, while the source arising from extensions is removed under Strong Warrant. 
Warrant and human consensus are nevertheless distinct constructs, so we expect the association to be positive but imperfect.

We independently elicited human judgments on the same 100 focal recommendation prompts using Prolific. Participants selected one of the two alternatives presented in each prompt. We recruited 63 participants over three waves and obtained a median of 21.5 judgments per prompt (range: 20--22). Participants were shown only the focal recommendation prompts and were not shown model responses, model confidence, transformed prompts, or certificate classifications.

Human consensus for each prompt was quantified as the normalized absolute difference
\[
c=\frac{|n_1-n_2|}{n_1+n_2},
\]
where $n_1$ and $n_2$ denote the number of participants selecting each alternative. The measure ranges from $0$, indicating an even split, to $1$, indicating unanimous agreement.

\subsubsection{Known-Groups Design.}
\label{design:known_groups}
To provide known-groups evidence of construct validity, we assembled 40
additional binary recommendation prompts organized into 10 four-question
sets. Each set contained one question designed to represent each of the four
warrant categories: No Warrant, Conditional Warrant, Basic Warrant, and
Strong Warrant. Five sets were independently developed by subject-matter
experts in their respective domains. We developed the remaining five sets and
obtained independent review from domain experts to verify their intended
warrant classifications. The expected classifications were established before
evaluating the target models. For the Conditional, Basic, and Strong Warrant cases, experts also identified the preferred alternative.
The known-groups scenarios span heterogeneous managerial and analytical decision settings, including hiring, hospitality, sourcing, and optimization.

We evaluated the known-groups prompts using \haiku, \sonnet, \nano, \mini, \sol, and \llamafour. The five non-Llama models were the same as in the primary evaluation. The two Llama models used in the primary analysis were no longer available in our evaluation environment when the known-groups study was conducted, so we used \llamafour\ as the Llama-family model.

\subsection{Validation Analyses Design}

\subsubsection{Content Validity.}

A classification was considered correct when the selected relationship matched the transformation's intended tier. Separately for each tier, we estimated the probability of a correct classification using logistic regression with two-way cluster-robust standard errors by participant and transformation item. We tested the one-sided hypothesis that this probability exceeded 0.5, requiring the intended tier to be selected more often than all alternative classifications combined. As a complementary item-level measure, we report the number of transformations for which a majority of participants selected the intended tier.

\subsubsection{Known-Groups Validity.}
\label{known_groups_validity}

We assessed whether assigned warrant increased according to the prespecified
ordering using Page's ordered-trend test. Each four-question set was treated as
a block, and the four expected warrant categories were treated as the
prespecified ordered conditions. The test therefore evaluates whether
model-assigned warrant increases systematically from No Warrant through Strong
Warrant while accounting for the blocked structure of the design. We
additionally report the mean assigned warrant within each known group.

As a complementary assessment of correspondence with the prespecified
categories, we computed the mean absolute error (MAE) between the
model-assigned and expert-designated warrant tiers. We evaluated the observed
MAE using a blocked permutation test in which the four expected warrant labels
were randomly permuted within each four-question set while model assignments
were held fixed. This procedure preserves the set structure while generating
the null distribution expected if model-assigned warrant were unrelated to the
prespecified tiers. We used 100,000 blocked permutations and additionally
report the percentage of prompts for which the assigned and prespecified
categories matched exactly.

Finally, we applied the Jonckheere--Terpstra ordered-trend test as a robustness
check. Unlike Page's test, the Jonckheere--Terpstra test treats the four known
groups as independent rather than accounting for the four-question-set blocks.
Because each group contained only 10 observations, we based inference on
permutation $p$-values. Agreement between the Page and Jonckheere--Terpstra tests would support the robustness of the ordered pattern to alternative treatments of the four-question set structure.

\subsubsection{Nomological Validity.}
\label{sec:nomological_design}

We assessed nomological validity by testing whether stronger warrant was
associated with greater human consensus on the underlying decision. For each
model, we computed Spearman's rank correlation between the four-level warrant
category and the human-consensus measure $c$.

We next examined the contributions of the individual certificate tiers in two
ways. First, we computed Spearman correlations between human consensus and
the violation rate at each tier. Second, we assessed the incremental
contribution of successive tiers using nested regressions that compared
progressively refined certificate classifications. Specifically, we tested
whether incorporating T1 improved explanatory power beyond T0 alone, whether
incorporating T2 improved explanatory power beyond the T0--T1 classification,
and whether incorporating T3 improved explanatory power beyond the T0--T2
classification. At each step, we report the increase in $R^2$ and test the
additional distinction introduced by the new tier using HC3
heteroskedasticity-robust standard errors~\citep{MacKinnonWhite1985}.

Because the 100 prompts were constructed from 22 base topics, we additionally examined whether the warrant--consensus relationship persisted within base topics. To test this, we estimated for each model a regression of human consensus on the warrant category, coded from 0 to 3, with base-topic fixed effects and standard errors clustered by base topic.

We further examined whether decision difficulty or related characteristics of the decision could account for the warrant--consensus relationship. Such characteristics may both reduce agreement among human decision-makers and make LLM recommendations less stable across the certificate's tests. Relevant factors include closer-valued alternatives, uncertainty about their values, ambiguity, unfamiliarity, and cognitive effort. Response time has been shown to reflect several of these characteristics and provides a broad behavioral measure of the demands involved in reaching a decision \citep{chabris2009allocation,lee2020valuecertainty,conte2023decisiontime,atzori2023responsetimes}. If the warrant--consensus relationship merely reflected prompts that were more difficult, ambiguous, or cognitively demanding, controlling for response time should substantially attenuate the warrant coefficient and leave little incremental variation explained by warrant. Our primary measure was median time to first response for each prompt. Because response times were right-skewed, we log-transformed this measure before estimation.

To assess whether warrant explained variation in human consensus beyond response time, we estimated
\begin{equation}
C_i
=
\beta_0
+
\beta_1 W_{im}
+
\beta_2 R_i
+
\varepsilon_{im},
\label{eq:incremental}
\end{equation}
where \(C_i\) denotes human consensus for prompt \(i\), \(W_{im}\) denotes the warrant category assigned by model \(m\), and \(R_i\) denotes log median time to first response. Human consensus, warrant, and log response time were standardized prior to estimation.

We compared the full model with a restricted model containing response time but not warrant and reported
\[
\Delta R^2_W
=
R^2_{R+W}
-
R^2_R,
\]
which measures the incremental variation in human consensus explained by warrant beyond response time. We also examined whether including response time attenuated the warrant coefficient relative to a warrant-only specification. If the warrant--consensus relationship primarily reflected decision difficulty or related characteristics of the decision, controlling for response time should substantially attenuate the warrant coefficient and leave little incremental variation explained by warrant. We used HC3 heteroskedasticity-robust standard errors throughout and repeated the analysis using log median total response time as a robustness check.

\subsubsection{Discriminant Validity.}

We assessed discriminant validity by examining whether epistemic warrant was empirically distinct from verbalized confidence, a theoretically related construct commonly used to characterize the support for LLM outputs. Confidence reflects the model's stated certainty in its recommendation, whereas warrant characterizes the strength and scope of support for that recommendation.

We examined this distinction in two ways. First, for each model, we computed Spearman's rank correlation between warrant category and verbalized confidence, with 95\% confidence intervals obtained from 10,000 bootstrap resamples of prompt-level pairs. Because the observed marginal distributions and ties restrict the maximum attainable Spearman correlation, we also computed the model-specific attainable maximum and used it to contextualize the observed association.

Second, we applied the incremental analysis in Equation~\ref{eq:incremental}, with verbalized confidence as $X$, to determine
whether warrant captured information about human consensus beyond that
contained in the model's stated confidence. Human consensus, verbalized
confidence, and warrant were rank-transformed before estimation. We report the
increase in $R^2$ from adding warrant to a confidence-only specification and
test the warrant coefficient using HC3 heteroskedasticity-robust standard
errors. Evidence that warrant was related to but not redundant with
confidence, and that adding warrant increased explained variation in human
consensus beyond confidence alone, would indicate that the two constructs were
empirically distinct.

\subsubsection{Robustness.}

Our primary analyses used the four-level categorical representation of warrant defined in Section~\ref{subsec:categories} and the original transformation-generation procedure. We assessed whether the nomological-validity results depended on either implementation choice.

First, we repeated the warrant--consensus analysis using two continuous
representations of certificate strength: a pooled score that aggregates
violations across all certificate queries and a lexicographic score that
preserves the tier hierarchy while retaining within-tier variation. Both
scores were reverse-coded so that higher values indicate stronger warrant.

Second, we regenerated the certificate queries using the alternative
transformation-generation procedure and recomputed warrant classifications
under the same four-level hierarchical rules for \haiku, \sonnet, \nano, \mini, \sol, and
\llamafour. We repeated the warrant--consensus analysis using the resulting classifications. 


\subsection{Data and Code Availability}
\label{sec:data_code}

All reproducibility materials are available at \url{https://osf.io/fzyk9}. The repository includes the 100 base recommendation prompts; all transformed prompts generated under both the primary and alternative robustness transformation procedures; the transformation instructions and implementation code; model responses, tier-level violations, warrant classifications, and verbalized-confidence responses; the human-consensus and indicator-validity data; and the complete known-groups prompt set with prespecified warrant classifications and an indication of whether each set was expert-generated or researcher-generated and independently expert-validated. The repository also contains the analysis code used to construct the certificates and reproduce all reported statistical tests and robustness analyses.

\section{Empirical Results}

This section  presents the empirical evidence for the validity of the reliance certificate. We first describe the distributions of warrant classifications and human consensus across the evaluation sample (Section~\ref{sec:descriptive_results}), then evaluate indicator content validity (Section~\ref{sec:indicator_validation_results}), known-groups validity (Section~\ref{sec:known_groups_results}), nomological validity (Section~\ref{sec:nomological_results}), discriminant validity (Section~\ref{sec:discriminant_results}), and the robustness of these results across alternative certificate representations (Section~\ref{sec:robustness_results}).

\subsection{Descriptive Results}\label{sec:descriptive_results}

{\OneAndAHalfSpacedXI
\begin{table}[ht]
\centering
\captionof{table}{Distribution of Warrant Categories Across Models ($n=100$)}
\label{tab:warrant_distribution}

{\footnotesize
\renewcommand{\arraystretch}{1.2}
\setlength{\tabcolsep}{6pt}

\begin{tabular}{
>{\raggedright\arraybackslash}p{0.14\textwidth}
>{\raggedright\arraybackslash}p{0.19\textwidth}
>{\centering\arraybackslash}p{0.12\textwidth}
>{\centering\arraybackslash}p{0.115\textwidth}
>{\centering\arraybackslash}p{0.115\textwidth}
>{\centering\arraybackslash}p{0.115\textwidth}
}
\hline
\textbf{Model Family} & \textbf{Model} & \textbf{No Warrant} & \textbf{Conditional} & \textbf{Basic} & \textbf{Strong} \\
\hline
\multirow{2}{*}{Claude}
& \haiku & 43 & 20 & 11 & 26 \\
& \sonnet & 30 & 25 & 14 & 31 \\
\hline
\multirow{3}{*}{GPT}
& \nano & 78 & 8 & 5 & 9 \\
& \mini & 47 & 17 & 9 & 27 \\
& \sol & 25 & 26 & 13 & 36 \\
\hline
\multirow{2}{*}{Llama}
& \llamafast & 81 & 3 & 1 & 15 \\
& \llama & 39 & 17 & 12 & 32 \\
\hline
\end{tabular}

\parbox{0.91\textwidth}{
\vspace{3pt}
\textit{Note.} Entries report the number of prompts assigned to each mutually exclusive warrant category. Each model was evaluated on the same 100 prompts.
}
}
\end{table}}

\paragraph{Warrant distribution across models.}
Table~\ref{tab:warrant_distribution} shows substantial variation in the
distribution of warrant classifications across models. The number of prompts
assigned No Warrant ranges from 25 for \sol\ to 81 for \llamafast, while the
number assigned Strong Warrant ranges from 9 for \nano\ to 36 for \sol. The
distribution also varies within each model family. For example, the number of
No Warrant classifications is 43 for \haiku\ and 30 for \sonnet, and 78, 47,
and 25 for \nano, \mini, and \sol, respectively. Within each family, the
distribution shifts monotonically toward higher warrant categories in the
capability ordering reported by the Artificial Analysis Intelligence
Index~\citep{ArtificialAnalysis2026}. In particular, both the number of Basic
Warrant assignments and the number of Strong Warrant assignments increase with the 
independently benchmarked model capability. 

\paragraph{Human consensus distribution.}
Human consensus also varies widely across the 100 prompts, with a median of
$0.67$ and an interquartile range of $[0.33,0.91]$. The observed values span
decisions on which participants were evenly divided through decisions on
which they were unanimous, providing variation across the full range of the
external criterion used for nomological validation.

\subsection{Content Validity} \label{sec:indicator_validation_results}

Human judgments strongly supported the intended tier assignments.
Participants classified 78.5\% of Tier~1 transformations, 97.9\% of Tier~2
transformations, and 72.2\% of Tier~3 transformations in accordance with their
intended tiers. In the logistic models accounting for repeated judgments by
both participants and transformation items, the estimated probability of a
correct classification exceeded 0.5 for all three tiers (all one-sided
$p<0.001$). The corresponding 95\% confidence intervals were
$[0.653,0.876]$ for Tier~1, $[0.928,0.994]$ for Tier~2, and
$[0.617,0.808]$ for Tier~3.

The item-level results were similarly consistent with the intended
classifications. A majority of participants selected the intended tier for
8 of 8 Tier~1 transformations, 8 of 8 Tier~2 transformations, and 7 of 8
Tier~3 transformations. Thus, the independently elicited judgments indicate
that the generated transformations generally instantiate the relationships
required by their intended certificate tiers.

\subsection{Known-Groups Validity}
\label{sec:known_groups_results}

The known-groups analysis provides  support for the
certificate's ability to distinguish among the four prespecified levels of
epistemic warrant. Among cases prespecified as Conditional, Basic, or Strong Warrant, the model’s focal recommendation agreed with the expert-identified alternative whenever its certificate assignment was above No Warrant. Mean assigned warrant was nondecreasing across the four
known groups for all six models. Page's ordered-trend test supported the
prespecified ordering for five models ($p\leq .035$), and strongly for three models ($p < 0.001$). The ordered trend was not statistically significant for \nano\ ($L=258.0$, $p=.208$); Nano assigned 36 of the 40 prompts No Warrant and the remaining four Strong Warrant, with no assignments to the Conditional or Basic categories. 
Overall,
the consistency of the ordered pattern across three model families indicates
that the certificate recovers theoretically meaningful differences in warrant.

\begin{table}[ht]
\centering
{\OneAndAHalfSpacedXI
\caption{Known-Groups Validation of Warrant}
\label{tab:known_groups}

{\footnotesize
\renewcommand{\arraystretch}{1.2}
\setlength{\tabcolsep}{6pt}

\begin{tabular}{
    >{\raggedright\arraybackslash}p{0.14\textwidth}
    >{\raggedright\arraybackslash}p{0.19\textwidth}
    >{\centering\arraybackslash}p{0.25\textwidth}
    >{\centering\arraybackslash}p{0.10\textwidth}
    >{\centering\arraybackslash}p{0.10\textwidth}
}
\hline
\multicolumn{1}{c}{\textbf{Model Family}} &
\multicolumn{1}{c}{\textbf{Model}} &
\multicolumn{1}{c}{\textbf{Group Means}} &
\multicolumn{1}{c}{\textbf{Page's $L$}} &
\multicolumn{1}{c}{\textbf{$p$}} \\
\hline

\multirow{2}{*}{Claude}
& \haiku  & [0.20, 0.40, 0.90, 1.70] & 267.5 & $0.035^{*}$ \\
& \sonnet & [0.30, 1.10, 1.80, 2.70] & 281.5 & $<0.001^{***}$ \\
\hline

\multirow{3}{*}{GPT}
& \nano & [0.00, 0.00, 0.60, 0.60] & 258.0 & $0.208$ \\
& \mini & [0.00, 0.20, 1.40, 2.40] & 280.0 & $<0.001^{***}$ \\
& \sol  & [0.30, 1.50, 2.20, 3.00] & 286.0 & $<0.001^{***}$ \\
\hline

Llama
& \llamafour & [0.30, 0.30, 0.70, 2.40] & 274.0 & $0.004^{**}$ \\
\hline
\end{tabular}

\parbox{0.78\textwidth}{
\vspace{3pt}
\textit{Note.} Group means are reported in the prespecified order: No Warrant,
Conditional Warrant, Basic Warrant, and Strong Warrant, coded 0, 1, 2, and 3,
respectively. Page's $L$ tests the prespecified ordered trend while treating
each four-question set as a block. $^{*}p<0.05$, $^{**}p<0.01$, and
$^{***}p<0.001$.
}
}
}
\end{table}

The additional known-groups analyses support the same conclusion. Permutation-based
Jonckheere--Terpstra tests support the ordered trend for the same five models
($p\leq .006$). Across all six models, assigned categories are  significantly
closer to the prespecified categories than expected under random within-set
reassignment: observed MAE ranges from 0.550 to 1.300, compared with mean null
MAE values ranging from 1.400 to 1.499 (all permutation-test $p$-values$\leq .043$). Exact
correspondence with the prespecified category ranges from 30.0\% for \nano\ to
60.0\% for \sol. Full results are reported in
Appendix~\ref{app:known_groups}. These analyses indicate that certificate-assigned warrant generally increases in the prespecified order and that the resulting assignments are closer to the intended categories than expected by chance, although discrimination is weaker for \nano.

\subsection{Nomological Validity}
\label{sec:nomological_results}


\begin{figure}[t]
\centering

\begin{subfigure}[t]{0.47\textwidth}
\centering
\includegraphics[width=\textwidth]{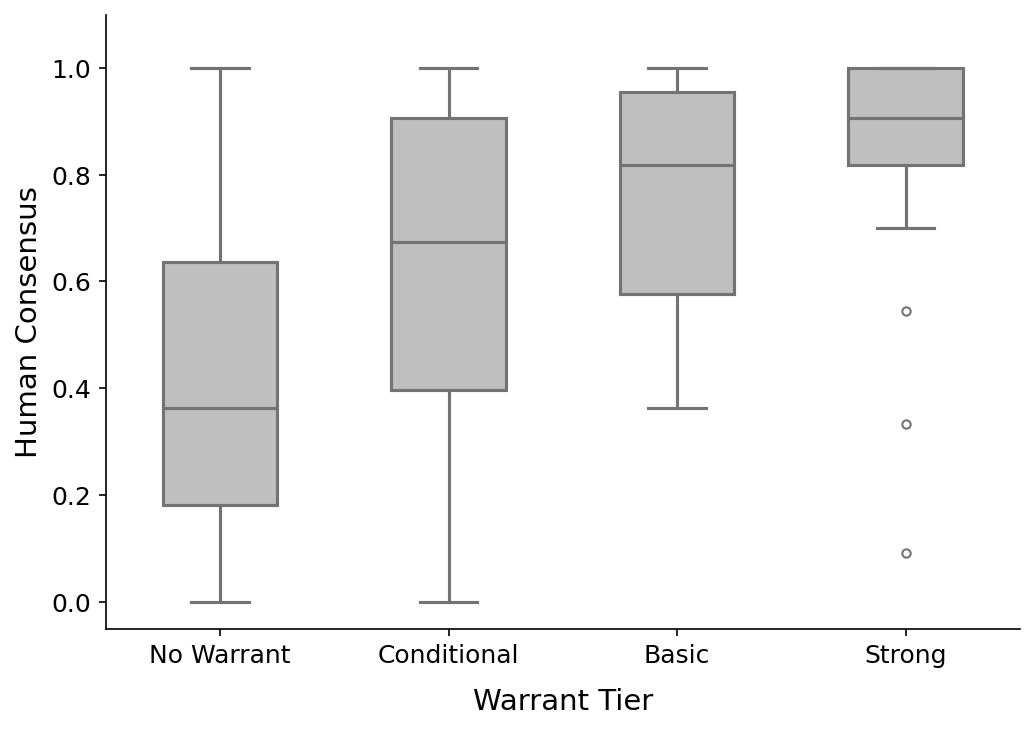}
\caption{\haiku}
\label{fig:warrant_boxplot_haiku}
\end{subfigure}
\hfill
\begin{subfigure}[t]{0.47\textwidth}
\centering
\includegraphics[width=\textwidth]{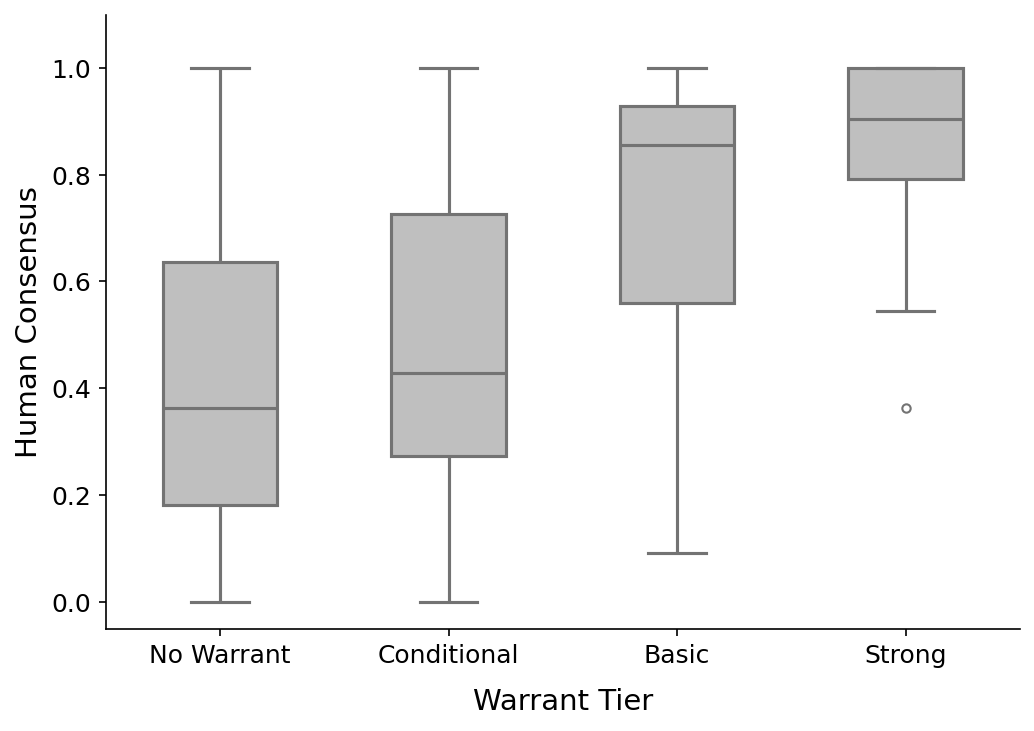}
\caption{\llama}
\label{fig:warrant_boxplot_llama}
\end{subfigure}

\caption{Human consensus by warrant category for \haiku\ and \llama.}
\label{fig:warrant_boxplot}
\end{figure}

\begin{table}[ht]
\centering
{\OneAndAHalfSpacedXI
\caption{Association Between Warrant and Human Consensus ($n=100$)}
\label{tab:nomological}

{\footnotesize
\renewcommand{\arraystretch}{1.2}
\setlength{\tabcolsep}{6pt}

\begin{tabular}{
>{\raggedright\arraybackslash}p{0.14\textwidth}
>{\raggedright\arraybackslash}p{0.19\textwidth}
>{\centering\arraybackslash}p{0.13\textwidth}
>{\centering\arraybackslash}p{0.13\textwidth}
}
\hline
\multicolumn{1}{c}{\textbf{Model Family}} &
\multicolumn{1}{c}{\textbf{Model}} &
\multicolumn{1}{c}{\textbf{Spearman $\rho$}} &
\multicolumn{1}{c}{\textbf{$p$}} \\
\hline
\multirow{2}{*}{Claude}
& \haiku  & 0.557 & $<0.001^{***}$ \\
& \sonnet & 0.497 & $<0.001^{***}$ \\
\hline
\multirow{3}{*}{GPT}
& \nano & 0.369 & $<0.001^{***}$ \\
& \mini & 0.520 & $<0.001^{***}$ \\
& \sol  & 0.411 & $<0.001^{***}$ \\
\hline
\multirow{2}{*}{Llama}
& \llamafast & 0.145 & $0.150$ \\
& \llama     & 0.585 & $<0.001^{***}$ \\
\hline
\end{tabular}

\parbox{0.68\textwidth}{
\vspace{3pt}
\textit{Note.} Spearman's $\rho$ measures the association between warrant category and human consensus across the 100 prompts. Higher warrant is expected to be associated with greater human consensus. $^{*}p<0.05$, $^{**}p<0.01$, and $^{***}p<0.001$.
}
}
}
\end{table}

As shown in Table~\ref{tab:nomological}, warrant is positively associated with
human consensus for all seven models, and the association is statistically
significant for six. Among these six models, Spearman correlations range from
$\rho=0.369$ for \nano\ to $\rho=0.585$ for \llama. The relationship is
weaker and not statistically significant for \llamafast\ ($\rho=0.145$,
$p=.150$). Figure~\ref{fig:warrant_boxplot} illustrates the relationship for
\haiku\ and \llama: prompts assigned stronger warrant by the model generally
elicit greater agreement among human respondents. Appendix~\ref{app:nomological}
reports the corresponding distributions for the remaining models.

The relationship also persists after accounting for the fact that multiple prompts were generated from each base topic. In base-topic fixed-effects models,
the warrant coefficient is positive and statistically significant for all
seven models, with coefficients ranging from 0.076 to 0.154
($p\leq .003$). Thus, the association is present not only across substantively
different topics but also among related prompts derived from the same base
topic. Full results are reported in
 Appendix~\ref{app:within_base_nomological}.

We next examine whether the warrant--consensus relationship is attributable
to a particular component of the certificate. Higher violation rates are
associated with lower human consensus at each tier. The association is
statistically significant for four models at T0, all seven models at T1, six
models at T2, and five models at T3
(Appendix~\ref{app:pertier}). The separate tier associations therefore
indicate that the overall relationship is not confined to a single type of
certificate query.

The nested regressions provide a stricter test of the incremental distinctions introduced by successive tiers.
 Incorporating T1 beyond T0 significantly
increases explained variation in human consensus for six of the seven models,
with $\Delta R^2$ ranging from 0.065 to 0.205 among those models.
Incorporating T2 beyond the T0--T1 classification provides a significant
additional contribution for four models, with $\Delta R^2$ ranging from
0.046 to 0.104. T3 produces smaller increments
($\Delta R^2=0.006$--$0.029$), none of which is statistically significant for
an individual model at the conventional 0.05 threshold. The smaller T3 increments should be interpreted in light of the certificate's hierarchical structure and the resulting category sizes. T3 affects only recommendations that have already passed T0--T2, and its incremental contribution comes specifically from distinguishing Basic from Strong Warrant. The Basic category contains only 1--14 observations across models, limiting both the number of classifications changed by T3 and the statistical power to detect an additional association with human consensus. Full model-specific results are reported in
Appendix~\ref{app:incremental}.

Finally, we assess whether decision difficulty or related characteristics of the decision could account for the warrant--consensus relationship. After controlling for log median time to first response, the standardized warrant coefficient remains positive and statistically significant for all seven models, ranging from 0.163 to 0.575. Relative to the warrant-only specifications, the coefficients decline by only approximately 3.5\%--12.4\%, and warrant contributes an incremental \(\Delta R^2\) of 0.026--0.312 beyond response time. In contrast, response time is statistically significant only for \llamafast\ and contributes comparatively little additional explanatory power (\(\Delta R^2=0.005\)--\(0.042\)). The results are substantively unchanged when using median total response time: warrant remains significant for every model, its coefficient declines by only 3.1\%--12.1\%, and total response time is not significant for any model (\(p\geq .052\)). Thus, the warrant--consensus relationship is not readily explained by decision difficulty or related characteristics reflected in response time. Full results are reported in Appendix~\ref{app:difficulty}.

\subsection{Discriminant Validity}
\label{sec:discriminant_results}

{\OneAndAHalfSpacedXI
\begin{table}[ht]
\centering
\captionof{table}{Spearman Correlation Between Warrant Category and Verbalized Confidence}
\label{tab:warrant_correlation}

{\footnotesize
\renewcommand{\arraystretch}{1.2}
\setlength{\tabcolsep}{6pt}

\begin{tabular}{
>{\raggedright\arraybackslash}p{0.14\textwidth}
>{\raggedright\arraybackslash}p{0.17\textwidth}
>{\centering\arraybackslash}p{0.10\textwidth}
>{\centering\arraybackslash}p{0.18\textwidth}
>{\centering\arraybackslash}p{0.10\textwidth}
}
\hline
\textbf{Model Family} & \textbf{Model} & \textbf{$\rho$} & \textbf{95\% CI} & \textbf{Cap} \\
\hline
\multirow{2}{*}{Claude}
& \haiku  & 0.641 & [0.516, 0.743] & 0.944 \\
& \sonnet & 0.591 & [0.451, 0.707] & 0.951 \\
\hline
\multirow{3}{*}{GPT}
& \nano & 0.227 & [0.048, 0.390] & 0.630 \\
& \mini & 0.465 & [0.288, 0.625] & 0.934 \\
& \sol  & 0.571 & [0.423, 0.692] & 0.957 \\
\hline
\multirow{2}{*}{Llama}
& \llamafast & 0.219 & [0.023, 0.406] & 0.713 \\
& \llama     & 0.541 & [0.378, 0.679] & 0.914 \\
\hline
\end{tabular}

\parbox{0.79\textwidth}{
\vspace{3pt}
\textit{Note.} $\rho$ denotes Spearman's rank correlation between warrant category and verbalized confidence. Confidence intervals are 95\% bootstrap intervals for $\rho$. Cap denotes the maximum attainable Spearman correlation given the observed marginal distributions and ties in the two variables.
}
}
\end{table}}

Warrant and verbalized confidence are positively related across all seven models (Table~\ref{tab:warrant_correlation}), with Spearman correlations ranging from 0.219 to 0.641. Because warrant contains only four ordered categories and both measures contain ties,  a correlation of one is not generally attainable. We therefore also compute the maximum Spearman correlation attainable under each model's observed marginal distributions. The observed correlations correspond to 30.7\%--67.9\% of these attainable maxima. The 95\% confidence intervals also remain well below the model-specific attainable maxima. Even the upper bound of each interval is below 80\% of the corresponding attainable correlation, with the largest normalized upper bound equal to 0.787.  Thus, although warrant and verbalized confidence are positively related, substantial variation in their rankings remains even after accounting for the mechanical restriction imposed by ties.

The distinction is also visible in individual recommendations. When asked
whether it is better for a music artist to sell 100 tickets to a 100-seat
venue or 750 tickets to a 1,000-seat venue, both \llama\ and \mini\ reported
100\% confidence in their recommendations, yet both failed the T0 and T1 tests
and received No Warrant. On the same decision, \sol\ reported 95\% confidence
but failed T1 and likewise received No Warrant. Conversely, when asked whether
Spanish or Russian is more useful to learn, \haiku\ reported only 45\%
confidence, yet its recommendation satisfied all certificate tiers and
received Strong Warrant. These examples illustrate that high confidence need
not imply strong warrant and that comparatively low confidence does not
preclude it.

More importantly, warrant explains variation in human consensus beyond that
captured by verbalized confidence. Adding warrant to a model containing
verbalized confidence increases explained variation for all seven models,
with $\Delta R^2$ ranging from 0.015 to 0.194. The warrant coefficient is
statistically significant for five models. The increment is not statistically
significant for \sol\ ($p=.063$) or \llamafast\ ($p=.210$), although it
remains positive for both. Full model-specific results are reported in
Appendix~\ref{app:discriminant}. Together, the incomplete correspondence
between warrant and confidence and the incremental explanatory value of
warrant provide complementary evidence that epistemic warrant captures
information distinct from models' stated confidence.

\subsection{Alternative Implementations}
\label{sec:robustness_results}

The results provide evidence that the warrant--consensus relationship is robust across models and alternative implementations of the certificate.

\paragraph{Alternative certificate operationalizations.}
Under the primary four-category measure, warrant is positively associated with
human consensus for all seven models and statistically significant for six.
The relationship is preserved when certificate strength is instead represented
using either of the two continuous measures. Under the lexicographic score,
Spearman correlations range from 0.287 to 0.620 and are statistically
significant for all seven models ($p\leq .004$). Under the pooled score,
correlations range from 0.199 to 0.540 and are likewise significant for every
model ($p\leq .047$). Both measures therefore recover a significant
relationship for \llamafast, for which the coarser four-category
classification produces a weaker and nonsignificant association. These
results show that the warrant--consensus relationship is not an artifact of
the categorical representation used in the primary analysis. Full results are
reported in Appendix~\ref{app:alternative_cert}.

\paragraph{Alternative transformation-generation procedure.}
The warrant--consensus relationship  persists when the certificate queries are regenerated using the alternative transformation-generation procedure. Across all six models evaluated under this procedure, warrant remains positively and significantly associated with human consensus, with Spearman correlations ranging from 0.278 to 0.443 (all \(p\leq .006\)). Thus, the observed warrant--consensus relationship is not specific to the transformation model, tier-specific instructions, or number of transformations used in the primary procedure. Full results are reported in Appendix~\ref{app:alternative_transformations}.


\section{Discussion}\label{sec:discussion}

This study develops epistemic warrant as a decision-level construct for characterizing the scope of support for a recommendation. We operationalize the construct for LLM recommendations through a four-tier reliance certificate. Following established guidance for construct validation~\citep{MacKenzie2011Construct}, we provide multiple forms of evidence for the validity of interpreting certificate assignments as measures of epistemic warrant. The findings suggest several theoretical and managerial implications.

\subsection{Theoretical Implications}

A central theoretical distinction of our framework is that the object of evaluation is an individual recommendation. Much of the existing literature on AI evaluation characterizes properties of models or systems, such as accuracy, robustness, calibration, or reliability. The epistemic warrant of a recommendation is not defined with respect to any specific model. The reliance certificate, by contrast, is model-dependent: the same recommendation may receive different certificate assignments depending on the model used to provide the behavioral evidence. Nevertheless, the underlying warrant remains a property of the recommendation.

Unlike much of the literature on robustness, warrant does not treat invariance
as an unconditional desideratum. A model may reasonably reverse a
recommendation across repeated generations when the alternatives are closely
balanced; in our framework, such variation provides a useful signal that the
model does not strongly prefer one alternative to the other. This suggests that not every observed inconsistency should be treated as a defect to be eliminated. When alternatives are substantively close, variation across repeated queries may reflect the absence of a strong basis for preferring one over the other, rather than a failure of the model.

The LLM setting makes it possible to operationalize epistemic conditions that are difficult to observe directly for human beliefs.  For human decision makers, questions concerning how a belief would behave under similar circumstances are difficult to instantiate without carryover effects. LLMs, however, can be queried in separate sessions without one query affecting the next, allowing systematically varied versions of the same decision to be examined independently. This makes it possible to obtain
behavioral evidence about the scope of support for a particular recommendation,
turning an abstract epistemic property into an empirically assessable
decision-level construct.

\subsection{Managerial Implications}

Epistemic warrant is a theoretical property of a recommendation; if it were directly observable, it would reveal the scope over which a recommendation is supported. The four categories differ qualitatively in where the boundary of support lies and therefore in what the decision maker needs to resolve before relying on the recommendation. No Warrant implies that the decision maker needs to establish whether there is a basis for relying on the recommendation at all; Conditional Warrant implies that the decision maker needs to resolve which refinement of the focal context applies to the decision at hand; Basic Warrant implies that the decision maker should determine whether the intended context might extend beyond the focal one; and Strong Warrant implies that reliance is broadly supported. Crucially, however, warrant cannot be observed directly. The reliance certificate therefore attempts to provide evidence about the underlying warrant using observed model behavior.

The certificate may imperfectly recover the underlying warrant for two reasons: incomplete coverage and model error. An individual certificate can examine only a finite set of probes. It is impossible to evaluate every semantically equivalent formulation and every reasonable refinement and extension. Insufficient coverage can only lead the certificate to overstate warrant: observing no reversal among the refinements examined, for example, does not establish that no reasonable refinement would produce a reversal. It is therefore important to choose transformations that provide broad and substantively relevant coverage, while recognizing that coverage cannot be exhaustive. In this paper, we deliberately use a generic transformation-generation procedure because it provides a standardized basis for evaluating heterogeneous decisions across topics and models. In applied settings, however, domain-specific knowledge could be used to generate refinements and extensions that better reflect the contingencies relevant to the decision at hand. Organizations could also allow decision makers to propose additional transformations reflecting contingencies they consider relevant, making the certificate more responsive to local knowledge that an automated generator may miss.

A second source of discrepancy is model error: observed model behavior may not perfectly track the underlying warrant. This can arise in two ways. First, response variability may cause an individual probe to yield an atypical response. The certificate deliberately uses variability at T0 and sensitivity to decision-preserving reformulations at T1 as evidence about support, but repeated evaluation of a given probe can help distinguish persistent behavior from an isolated response~\citep{yin2025fragile}. Second, the model may systematically evaluate the decision incorrectly for a variety of reasons, including knowledge limitations, reasoning errors, spurious associations or intrinsic  bias. In such cases, repeated probing will not eliminate the discrepancy.  A model may be ``stubborn" and assign a Strong Warrant to every recommendation. Before deployment, organizations could assess candidate models across diverse recommendations to identify models that assign uniformly high or low certificate categories despite substantive differences among the decision problems. Our known-groups study provides some evidence that model capability may reduce such discrepancies. Within each model family, correspondence between certificate assignments and the prespecified warrant ordering increases monotonically with independently benchmarked model capability.\footnote{Capability as reported by the Artificial Analysis Intelligence Index~\citep{ArtificialAnalysis2026}.} This pattern suggests that more capable models may provide behavioral evidence that more closely tracks underlying warrant, although our study does not identify whether this results from fewer anomalous responses, fewer systematic judgment errors, or both.

Despite these limitations, the reliance certificate can serve as a lightweight layer between an LLM recommendation and action. In many decision settings, verifying every recommendation may be costly or impractical. A large literature on selective prediction and learning develops mechanisms for determining which cases should be allocated to human decision makers for verification \citep{Madras2018Predict,Mozannar2020Defer}. The reliance certificate addresses a complementary question: not only \emph{whether} additional scrutiny is warranted, but \emph{what that scrutiny should focus on}. Depending on the warrant category, the unresolved issue may concern the basis for the focal recommendation, which refinement of the focal context applies, or whether the recommendation is being extended beyond the context in which it is supported.

The warrant categories capture qualitative differences in the scope of support; once a violation occurs at a given tier, additional violations at that tier do not alter the qualitative classification. The difference between Conditional and Basic Warrant, for instance, is whether the focal context must be further resolved into subcontexts, not how many such subcontexts produce reversals. Nevertheless, differences in violation rates may be quantitatively informative. Furthermore, the particular transformations under which the response reverses can be useful diagnostically by directing the decision maker's attention toward contingencies that may limit the scope of support. The certificate retains the number of probes and violations at each tier, allowing the categorical assignment to be supplemented with this information. Deployed systems could also generate additional probes around observed reversals to characterize these boundaries more precisely. Exposing the underlying probes and responses also makes the warrant assessment more transparent by showing the evidence on which the classification is based without purporting to explain the model's internal reasoning \citep{Rai2020Explainable,deFineLicht2020Transparency}.

The certificate does not by itself determine whether an organization should act on a recommendation. The same certificate may provide sufficient evidence for action in a low-stakes, reversible decision but warrant additional scrutiny when the stakes are high or the decision is difficult to reverse. Verification costs and access to external expertise may also affect the appropriate response. The certificate should therefore be treated as one input into the reliance decision rather than as a standalone action rule.

\subsection{Boundary Conditions and Future Research}

Several boundaries of the present framework point to directions for future research. First, future work could examine how reliance certificates should be designed under practical resource constraints. Certificate design involves both which contexts to examine and how intensively to probe them. Broader probing can improve contextual coverage, while repeated evaluation can reduce the influence of anomalous responses; both incur additional token, latency, and computational costs. Future research could therefore examine how to identify the most informative probes, allocate evaluation effort across and within contexts, and develop adaptive allocation or stopping rules that concentrate evaluation where additional information is most valuable.

Second, future research should examine how warrant information, and the certificate evidence used to assess it, should support reliance decisions and under what conditions different uses are most effective. This includes determining when warrant should support direct reliance, trigger additional verification or human review, or redirect attention toward unresolved contextual conditions, as well as what probe-level information should accompany the warrant category. The appropriate use may depend on factors such as the stakes and reversibility of the decision, verification costs, and access to external expertise.

A central challenge for future research is determining how reliance-certificate information should best be incorporated into decision making. Possible uses include supporting direct reliance, triggering additional verification or human review, and directing attention toward unresolved contextual conditions or potential boundaries of support. Prior research commonly evaluates appropriate reliance in terms of accepting correct AI advice while rejecting incorrect advice \citep{schemmer2023appropriate}, which requires observable ground truth, realized outcomes, or another independent benchmark \citep{kawakami2023training}. Comparing alternative uses of the certificate is therefore particularly challenging because the settings in which epistemic warrant is most useful are often precisely those in which ground truth is unavailable, delayed, or contested, while the underlying warrant itself is also unobservable. Future research should therefore examine which forms of certificate-informed decision support perform best, for whom, and under what conditions, using designs in which decision quality can ultimately be assessed against an independent criterion.


Third, our empirical implementation focuses on pairwise recommendations. Many organizational decisions involve larger choice sets, rankings, sequential decisions, or recommendations accompanied by quantitative predictions, although some larger-choice and ranking problems can themselves be implemented through sequences of pairwise comparisons~\citep{qin2023large}. Extending epistemic warrant to these settings may require different notions of reversal and different certificate structures, but the underlying idea of characterizing the scope over which a recommendation is supported may extend beyond pairwise choice.

\OneAndAHalfSpacedXI
\bibliographystyle{informs2014}
\bibliography{refs}
\newpage

\begin{APPENDICES}

\section{Illustrative Certificate Outcomes}
\label{app:examples}

Table~\ref{tab:certificate_examples} provides illustrative examples of focal prompts, their human-consensus values, and the resulting warrant classifications across the seven focal models. The examples were selected to illustrate the range of empirical patterns in the data, including cases in which warrant and human consensus align closely and cases in which they diverge. They are included for illustration only; all validation analyses use the full set of 100 prompts.

{\OneAndAHalfSpacedXI
\begin{table}[ht]
\centering
\caption{Illustrative Warrant Classifications Across Models}
\label{tab:certificate_examples}

{\scriptsize
\renewcommand{\arraystretch}{1.25}
\setlength{\tabcolsep}{3pt}

\begin{tabular}{
>{\raggedright\arraybackslash}p{0.29\textwidth}
>{\centering\arraybackslash}p{0.08\textwidth}
>{\centering\arraybackslash}p{0.07\textwidth}
>{\centering\arraybackslash}p{0.07\textwidth}
>{\centering\arraybackslash}p{0.08\textwidth}
>{\centering\arraybackslash}p{0.08\textwidth}
>{\centering\arraybackslash}p{0.07\textwidth}
>{\centering\arraybackslash}p{0.07\textwidth}
>{\centering\arraybackslash}p{0.07\textwidth}
}
\hline
& \textbf{Human} &
\multicolumn{2}{c}{\textbf{Claude}} &
\multicolumn{3}{c}{\textbf{GPT}} &
\multicolumn{2}{c}{\textbf{Llama}} \\
\cmidrule(lr){3-4}
\cmidrule(lr){5-7}
\cmidrule(lr){8-9}
\textbf{Focal prompt} &
\textbf{Consensus} &
\textbf{haiku} &
\textbf{sonnet} &
\textbf{4.1-nano} &
\textbf{5.4-mini} &
\textbf{5.6-sol} &
\textbf{3.1-8b} &
\textbf{3.3-70b} \\
\hline

Which job would be easier for an AI to perform, software developer or financial analyst?
& 0.000 & None & None & None & None & None & None & None \\

Which is harder: winning a Nobel Prize or becoming an astronaut?
& 0.000 & None & Cond. & None & None & None & None & None \\

Which is more energy efficient per person, an electric car or train?
& 0.091 & Strong & Strong & None & Basic & Basic & None & Basic \\

Which energy source has a greater environmental impact for the same amount of energy, coal or solar?
& 0.273 & Cond. & Cond. & Basic & Basic & Basic & Basic & Basic \\

When estimating total project cost, is it more important to consider labor costs or schedule delays?
& 0.636 & None & Cond. & Basic & Strong & None & None & Basic \\

Which hotels are usually nicer, hotels near the airport or hotels in a resort area?
& 1.000 & Basic & Strong & Basic & Strong & Strong & None & Basic \\

Which is safer to travel by, motorcycle or train?
& 1.000 & Strong & Strong & Strong & Strong & Strong & Strong & Strong \\

\hline
\end{tabular}

\parbox{0.96\textwidth}{
\vspace{3pt}
\textit{Note.} Human consensus ranges from 0 (an even split) to 1 (unanimous agreement). Warrant classifications are abbreviated as None = No Warrant and Cond. = Conditional; Basic and Strong denote Basic Warrant and Strong Warrant, respectively.
}
}
\end{table}}



\section{Certificate Transformation Procedure}
\label{app:certificate_implementation}

All transformations used in the primary procedure were generated using \haiku. The prompt blocks below reproduce the instructions governing the content of each transformation, with values inserted programmatically represented by bracketed placeholders. For readability, we omit instructions specifying machine-readable output formats and other parsing requirements. The complete executable prompts and generation code are available in the replication repository.

\subsection*{Tier 0: Stochastic Stability}

Tier~0 does not require a generated transformation. The unchanged focal prompt is queried repeatedly to assess whether the recommendation is stable under repeated sampling.

\subsection*{Tier 1: Invariance Transformations}

Tier~1 evaluates whether the recommendation is invariant to transformations that preserve the underlying decision. We use option permutation, valence inversion, and semantically close contextual substitutions. We additionally construct a transformation combining option permutation and valence inversion.

\paragraph{Option permutation.}\mbox{}\par
\medskip 

\begingroup
\centering
{\scriptsize \ttfamily
\begin{tabular}{p{0.94\textwidth}}
\hline
Given this prompt: ``[FOCAL PROMPT]'' \\[4pt]
Swap the positions of the two options ``[OPTION A]'' and ``[OPTION B]'' so that wherever ``[OPTION A]'' appears (or is referred to) it is replaced by ``[OPTION B]'', and vice versa. The rest of the prompt must remain identical---do not change any other words, punctuation, or phrasing. Important: options may share a trailing noun (e.g., ``best-case or worst-case estimate''---here both options share ``estimate''). Handle these gracefully so the result reads naturally. \\[2pt]
\hline
\end{tabular}
}
\par
\endgroup

\medskip 
\paragraph{Valence inversion.}\mbox{}\par
\medskip 

\begingroup
\centering
{\scriptsize \ttfamily
\begin{tabular}{p{0.94\textwidth}}
\hline
Given this prompt: ``[FOCAL PROMPT]'' \\[4pt]
Invert its valence such that a respondent who would answer ``[OPTION A]'' to the original would answer ``[OPTION B]'' to the rephrased version, and vice versa. You can do this by, for example, replacing the comparison word with its antonym (e.g., ``better'' $\rightarrow$ ``WORSE'', ``prefer/choose'' $\rightarrow$ ``AVOID'', ``best'' $\rightarrow$ ``WORST'') or reframing the criterion (e.g., ``which is safer'' $\rightarrow$ ``which is RISKIER''). Change ONLY ONE word or short phrase---do not apply multiple inversions at once. Keep ``[OPTION A]'' and ``[OPTION B]'' in their original positions and unchanged. Also add a short explaining sentence, while ensuring that the answer to the rephrased version is reversed relative to the original. Do not add any additional assumptions or context. \\[2pt]
\hline
\end{tabular}
}
\par
\endgroup

\medskip 

The combined transformation is produced by applying the option-permutation procedure to the valence-inverted prompt.

\paragraph{Contextual substitution.}\mbox{}\par
\medskip 

\begingroup
\centering
{\scriptsize \ttfamily
\begin{tabular}{p{0.94\textwidth}}
\hline
Given this prompt: ``[FOCAL PROMPT]'' \\[4pt]
Replace ONLY one word in the context ``[CONTEXT]'' with a very close synonym---the meaning should be almost identical. Return the substituted words as simple root words without possessives, plurals, or other inflections (e.g., ``kid'' not ``kid's'', ``room'' not ``rooms''). \\[2pt]
\hline
\end{tabular}
}
\par
\endgroup

\medskip 

Each proposed substitution was evaluated using GloVe cosine distance, represented on a scale from 0 to 200, with smaller values indicating greater semantic similarity. Substitutions with distance no greater than 35 were retained as Tier~1 transformations. When a proposed substitution did not satisfy this criterion or could not be validated, the transformation model was instructed to generate a different substitution. More distant substitutions generated during this procedure were not used in the final certificate.

\subsection*{Tier 2: Narrower-Context Transformations}

Tier~2 evaluates whether the recommendation is preserved when the focal context is narrowed.

\paragraph{Context narrowing.}\mbox{}\par
\medskip 

\begingroup
\centering
{\scriptsize \ttfamily
\begin{tabular}{p{0.94\textwidth}}
\hline
Given this prompt: ``[FOCAL PROMPT]'' \\[4pt]
The context word is: ``[CONTEXT]''. Replace the context term with a MORE SPECIFIC term that narrows the scope of the question. In other words, the new context should be a particular example or instance of the original context. You should think of this as adding a constraint to the original context to narrow down its scope. Keep ``[OPTION A]'' and ``[OPTION B]'' unchanged. \\[2pt]
\hline
\end{tabular}
}
\par
\endgroup

\medskip 

\medskip 

Eight Tier~2 transformations were generated for each focal prompt within the same conversation by requesting different substitutions from those previously generated. All eight generated transformations were used in the certificate.

\subsection*{Tier 3: Broader Context Transformations}

Tier~3 evaluates whether the recommendation generalizes beyond the focal context, either to a broader context or to a related but non-nested context. The transformation prompt used to generate Tier~3 candidates targeted broader contexts.

\paragraph{Context expansion.}\mbox{}\par
\medskip 

\begingroup
\centering
{\scriptsize \ttfamily
\begin{tabular}{p{0.94\textwidth}}
\hline
Given this prompt: ``[FOCAL PROMPT]'' \\[4pt]
The context word is: ``[CONTEXT]''. Replace the context term with a MORE GENERAL term that broadens the scope of the question. You can think of this as removing a constraint to make the scope more broad or more open-ended. The new context should be a broader category or class that the original context falls under. Keep ``[OPTION A]'' and ``[OPTION B]'' unchanged. \\[2pt]
\hline
\end{tabular}
}
\par
\endgroup

\medskip 

\medskip 

Eight Tier~3 transformations were generated for each focal prompt within the same conversation by requesting different substitutions from those previously generated. All eight generated transformations were used in the certificate. Although the generation prompt targeted broader contexts, some generated transformations instead represented related but non-nested contexts, which also satisfy the predefined Tier~3 definition. The extent to which the generated transformations instantiated their intended tier relationships was assessed independently in the indicator-validity experiment described in Section~\ref{design:indicator}.

\subsection{Alternative Transformation-Generation Procedure}
\label{app:alternative_procedure}

As a robustness check, we regenerated the certificate probes using a revised transformation-generation procedure. The alternative procedure preserved the same four-tier structure and certificate rules but modified how the transformations themselves were generated. For T2 and T3, the generator was instructed to produce a more specific or more general version of the focal context without being explicitly supplied with the context term. For T1 paraphrasing, we replaced the original word-substitution procedure and GloVe-distance filter with five full-prompt paraphrases instructed to preserve the meaning as closely as possible while leaving the two alternatives unchanged. We also revised the permutation and valence-inversion instructions to better accommodate heterogeneous prompt formats. These changes provide a substantively different implementation of the probe-generation stage while leaving the interpretation of tier violations and warrant categories unchanged. The complete code appears in the replication repository.

\section{Indicator-Validity Experiment}
\label{app:indicator_prompts}

Participants were shown pairs consisting of a focal recommendation prompt and one generated transformation, labeled Question~1 and Question~2, and were asked to characterize the relationship between the two situations. They were not shown model responses, warrant classifications, or the tier for which the transformation had been generated.

The response options were:

\begin{enumerate}
    \item Question~1 describes a narrower instance of the situation in Question~2.
    \item Question~2 describes a narrower instance of the situation in Question~1.
    \item The two questions describe situations of approximately equal breadth.
    \item The situations are related, but neither is a broader version of the other.
    \item I cannot determine the relationship.
\end{enumerate}

Responses were mapped to the intended transformation tiers according to the position of the focal and transformed prompts in the pair. Approximately equal breadth corresponds to Tier~1; a transformed prompt that is narrower than the focal prompt corresponds to Tier~2; and a transformed prompt that is broader than the focal prompt, or related but non-nested, corresponds to Tier~3.

\section{Additional Results for Known-Groups Validity}
\label{app:known_groups}

As a supplementary known-groups analysis, we assess the distance between the model-assigned and expert-designated warrant tiers using mean absolute error (MAE). To preserve the set-specific block structure, we construct the null distribution by independently permuting the four expected warrant labels within each expert's four-question set while holding the model-assigned warrant values fixed. Table~\ref{tab:known_groups_mae} reports the observed MAE, the mean MAE under the permutation distribution, the corresponding permutation $p$-value, and the percentage of prompts for which the assigned and expert-designated tiers match exactly.
\begin{table}[ht]
\centering
{\OneAndAHalfSpacedXI
\caption{Supplementary Known-Groups Validation Results}
\label{tab:known_groups_mae}

{\footnotesize
\renewcommand{\arraystretch}{1.2}
\setlength{\tabcolsep}{6pt}
\begin{tabular}{
    >{\raggedright\arraybackslash}p{0.14\textwidth}
    >{\raggedright\arraybackslash}p{0.19\textwidth}
    >{\centering\arraybackslash}p{0.11\textwidth}
    >{\centering\arraybackslash}p{0.12\textwidth}
    >{\centering\arraybackslash}p{0.12\textwidth}
    >{\centering\arraybackslash}p{0.12\textwidth}
}
\hline
\multicolumn{1}{c}{\textbf{Model Family}} &
\multicolumn{1}{c}{\textbf{Model}} &
\multicolumn{1}{c}{\textbf{MAE}} &
\multicolumn{1}{c}{\textbf{Null MAE}} &
\multicolumn{1}{c}{\textbf{$p$}} &
\multicolumn{1}{c}{\textbf{Exact Match}} \\
\hline

\multirow{2}{*}{Claude}
& \haiku  & 1.000 & 1.424 & $0.003^{**}$ & 37.5\% \\
& \sonnet & 0.675 & 1.425 & $<0.001^{***}$ & 55.0\% \\
\hline

\multirow{3}{*}{GPT}
& \nano & 1.300 & 1.499 & $0.043^{*}$ & 30.0\% \\
& \mini & 0.700 & 1.462 & $<0.001^{***}$ & 52.5\% \\
& \sol  & 0.550 & 1.400 & $<0.001^{***}$ & 60.0\% \\
\hline

Llama
& \llamafour & 0.875 & 1.425 & $<0.001^{***}$ & 40.0\% \\
\hline
\end{tabular}

\parbox{0.82\textwidth}{
\vspace{3pt}
\textit{Note.} MAE is the mean absolute difference between model-assigned and expert-designated warrant tiers, coding No Warrant = 0, Conditional Warrant = 1, Basic Warrant = 2, and Strong Warrant = 3. Null MAE is the mean MAE obtained under random within-set permutation of the four expected warrant labels. Reported $p$-values are based on 100{,}000 blocked permutations. Exact Match is the percentage of the 40 known-groups prompts for which the model-assigned warrant tier exactly matched the expert-designated tier. $^{*}p<0.05$, $^{**}p<0.01$, and $^{***}p<0.001$.
}
}
}
\end{table}

As a robustness check, we repeated the ordered-trend analysis using the
Jonckheere--Terpstra test. Unlike Page's test, which accounts for the
four-question-set blocks, the Jonckheere--Terpstra test treats the four
known groups as independent. As shown in Table~\ref{tab:known_groups_jt},
the test supports the prespecified increasing ordering for five of the six
models ($p\leq .0058$). The trend does not reach the conventional
significance threshold for \nano\ ($J=340.0$, permutation $p=.0504$).
Thus, the Jonckheere--Terpstra results reproduce the model-level pattern
obtained with Page's test, indicating that the principal known-groups
conclusion does not depend on accounting for the set-level block structure.
\begin{table}[ht]
\centering
\caption{Jonckheere--Terpstra Robustness Test for the Known-Groups Ordering}
\label{tab:known_groups_jt}

{\footnotesize
\renewcommand{\arraystretch}{1.2}
\setlength{\tabcolsep}{5pt}

\begin{tabular}{
>{\raggedright\arraybackslash}p{0.14\textwidth}
>{\raggedright\arraybackslash}p{0.25\textwidth}
>{\centering\arraybackslash}p{0.13\textwidth}
>{\centering\arraybackslash}p{0.18\textwidth}
}
\hline
\multicolumn{1}{c}{\textbf{Model Family}} &
\multicolumn{1}{c}{\textbf{Model}} &
\multicolumn{1}{c}{\textbf{$J$}} &
\multicolumn{1}{c}{\textbf{Permutation $p$}} \\
\hline
\multirow{2}{*}{Claude}
    & \haiku  & 387.5 & $0.0058^{**}$ \\
    & \sonnet & 458.0 & $<0.0001^{***}$ \\
\hline
\multirow{3}{*}{GPT}
    & \nano & 340.0 & $0.0504$ \\
    & \mini & 452.0 & $<0.0001^{***}$ \\
    & \sol  & 487.5 & $<0.0001^{***}$ \\
\hline
Llama
    & \llamafour & 410.0 & $0.0011^{**}$ \\
\hline
\end{tabular}

\parbox{0.76\textwidth}{
\vspace{3pt}
\textit{Note.} The Jonckheere--Terpstra test evaluates the prespecified
increasing ordering from No Warrant through Strong Warrant while treating
the four known groups as independent. Reported $p$-values are based on
one-sided permutation tests. Each known group contains 10 observations.
$^{*}p<0.05$, $^{**}p<0.01$, and $^{***}p<0.001$.
}
}
\end{table}

\section{Additional Results for Nomological Validity}
\label{app:nomological}

\subsection{Warrant and Human Consensus Across Models}\label{app:warrant_consensus}

\begin{figure}[t]
\centering
\begin{subfigure}[t]{0.3\textwidth}
\centering
\includegraphics[width=\textwidth]{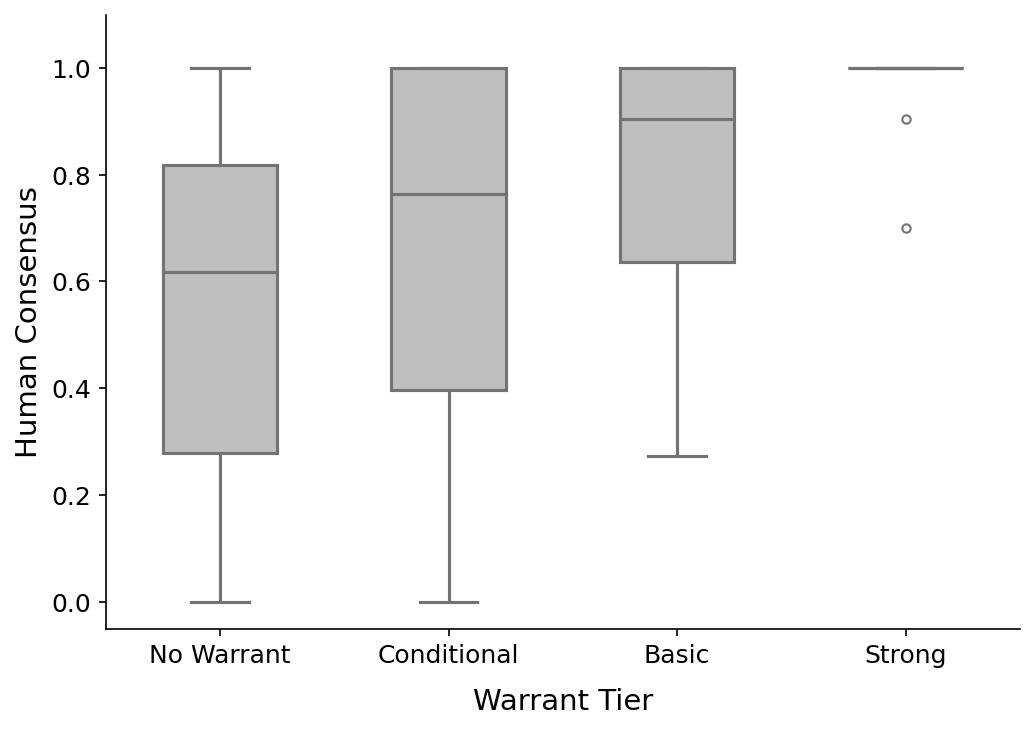}
\caption{\nano}
\label{fig:warrant_boxplot_a}
\end{subfigure}
\hfill
\begin{subfigure}[t]{0.3\textwidth}
\centering
\includegraphics[width=\textwidth]{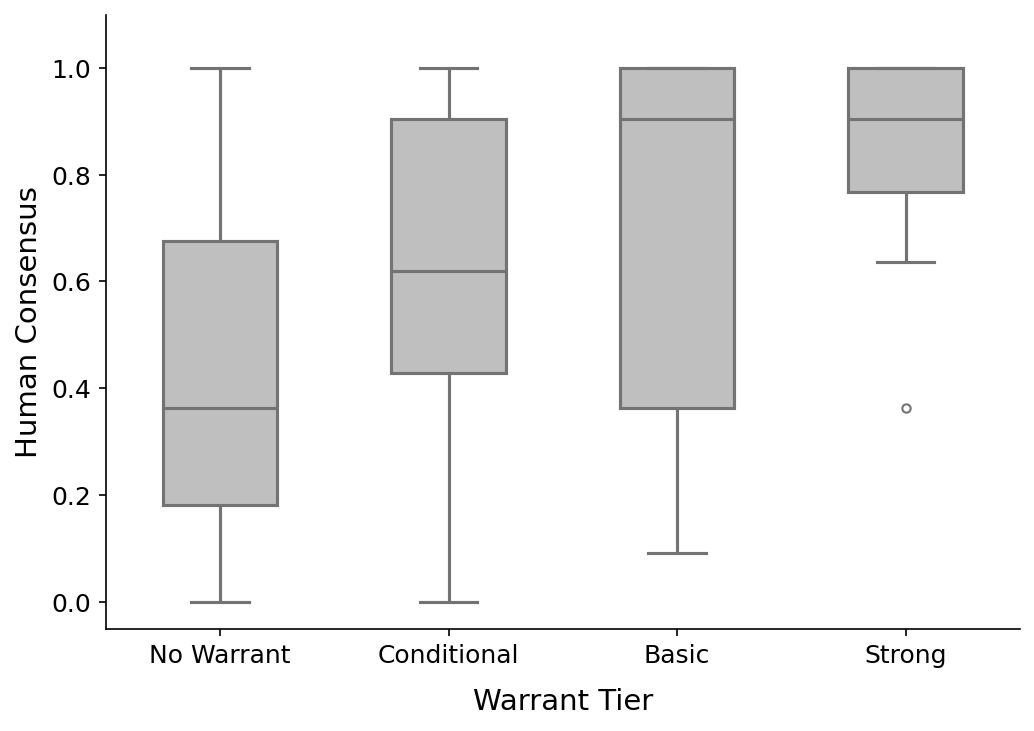}
\caption{\mini}
\label{fig:warrant_boxplot_b}
\end{subfigure}
\hfill
\begin{subfigure}[t]{0.3\textwidth}
\centering
\includegraphics[width=\textwidth]{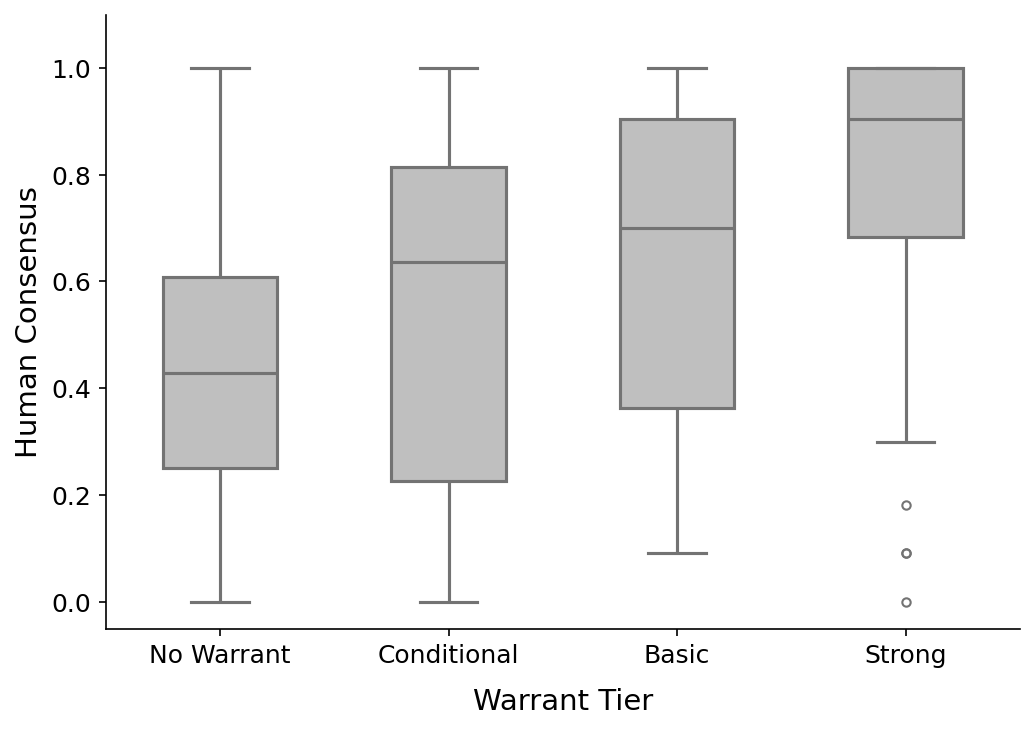}
\caption{\sol}
\label{fig:warrant_boxplot_c}
\end{subfigure}

\vspace{1em}

\begin{subfigure}[t]{0.3\textwidth}
\centering
\includegraphics[width=\textwidth]{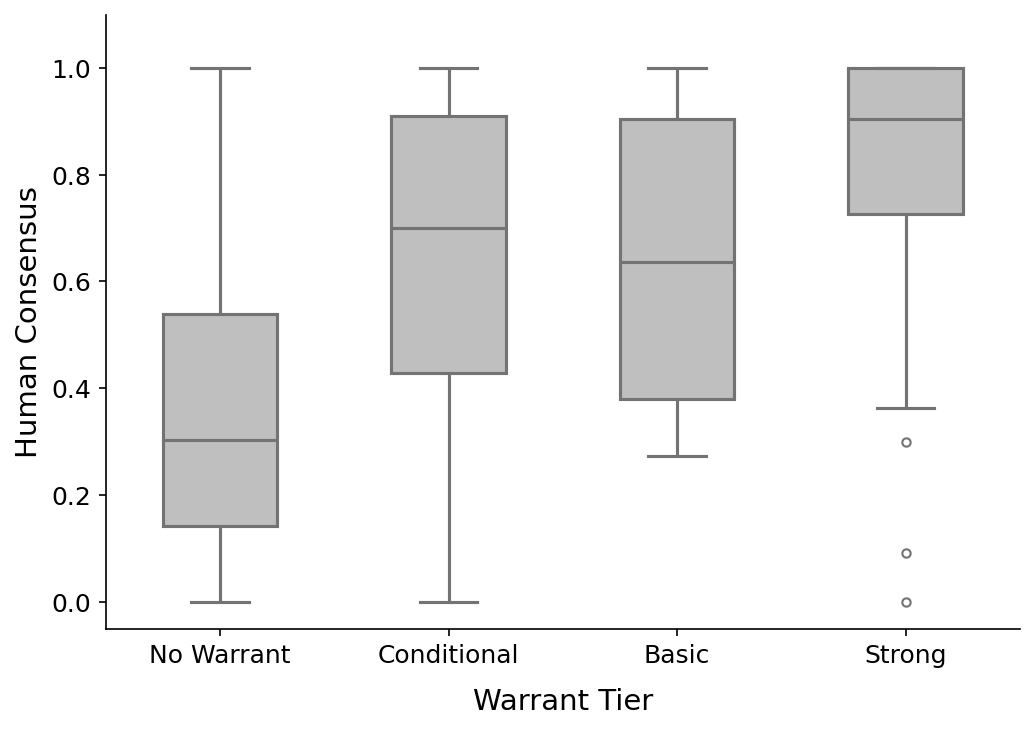}
\caption{\sonnet}
\label{fig:warrant_boxplot_d}
\end{subfigure}
\hspace{0.02\textwidth}
\begin{subfigure}[t]{0.3\textwidth}
\centering
\includegraphics[width=\textwidth]{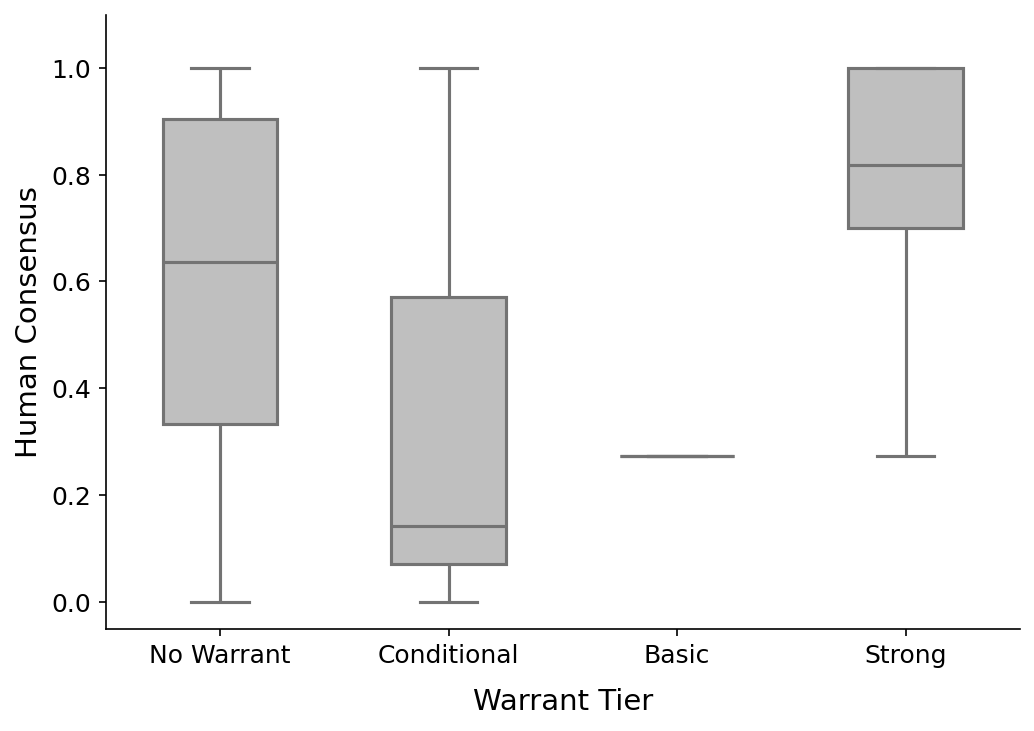}
\caption{\llamafast}
\label{fig:warrant_boxplot_e}
\end{subfigure}

\caption{Human consensus by warrant category for the five models not displayed in the main text.}
\label{fig:warrant_boxplot_app}
\end{figure}

Figure~\ref{fig:warrant_boxplot_app} shows the relationship between warrant
category and human consensus for the five models not displayed in the main
text. For \nano, \mini, and \sol, human consensus generally increases across
warrant categories, consistent with the patterns shown for \haiku\ and
\llama\ in the main text. For \sonnet, the overall relationship is positive
but not strictly monotonic: median consensus in the Basic Warrant category is
slightly below that of the Conditional category. For \llamafast, the warrant
categories are less clearly separated, consistent with the weaker association
between categorical warrant and human consensus observed for this model.

\subsection{Within-Base Robustness of the Nomological Validity Test}
\label{app:within_base_nomological}

Because the 100 derived questions are nested within 22 base topics,
observations sharing the same base topic may not be fully independent. We
therefore complement the primary Spearman analysis with base-topic
fixed-effects regressions in which human consensus is regressed on warrant
rank, with standard errors clustered by base topic.
Table~\ref{tab:within_base_consensus} reports the resulting coefficients,
clustered standard errors, and $p$-values for each model.

\begin{table}[ht]
\centering
{\OneAndAHalfSpacedXI
\caption{Base-Topic Fixed-Effects Robustness Test for the Association Between Warrant and Human Consensus}
\label{tab:within_base_consensus}

{\footnotesize
\renewcommand{\arraystretch}{1.2}
\setlength{\tabcolsep}{6pt}

\begin{tabular}{
>{\raggedright\arraybackslash}p{0.14\textwidth}
>{\raggedright\arraybackslash}p{0.19\textwidth}
>{\centering\arraybackslash}p{0.11\textwidth}
>{\centering\arraybackslash}p{0.13\textwidth}
>{\centering\arraybackslash}p{0.10\textwidth}
}
\hline
\multicolumn{1}{c}{\textbf{Model Family}} &
\multicolumn{1}{c}{\textbf{Model}} &
\multicolumn{1}{c}{\boldmath$\beta$} &
\multicolumn{1}{c}{\textbf{Clustered SE}} &
\multicolumn{1}{c}{\textbf{$p$}} \\
\hline
\multirow{2}{*}{Claude}
& \haiku  & 0.150 & 0.034 & $<0.001^{***}$ \\
& \sonnet & 0.120 & 0.039 & $0.002^{**}$ \\
\hline
\multirow{3}{*}{GPT}
& \nano & 0.154 & 0.036 & $<0.001^{***}$ \\
& \mini & 0.131 & 0.033 & $<0.001^{***}$ \\
& \sol  & 0.106 & 0.034 & $0.002^{**}$ \\
\hline
\multirow{2}{*}{Llama}
& \llamafast & 0.076 & 0.021 & $<0.001^{***}$ \\
& \llama     & 0.154 & 0.032 & $<0.001^{***}$ \\
\hline
\end{tabular}

\parbox{0.76\textwidth}{
\vspace{3pt}
\textit{Note.} The dependent variable is human consensus. Warrant is coded from 0 (No Warrant) to 3 (Strong Warrant). Each specification includes base-question fixed effects, and standard errors are clustered by base topic. All specifications use $n=100$ prompts nested within 22 base-question groups. $^{*}p<0.05$, $^{**}p<0.01$, and $^{***}p<0.001$.
}
}
}
\end{table}

\subsection{Associations by Certificate Tier}\label{app:pertier}

Table~\ref{tab:pertier_correlation} examines the association between human
consensus and violation rates at each individual certificate tier. T1
violation rates are significantly associated with human consensus for all
seven models, while T2 violation rates are significantly associated with
consensus for six of the seven models. T3 violation rates are significantly
associated with consensus for five models. T0 shows greater variation across
models, with significant associations for four models. These results indicate
that the overall warrant--consensus relationship is not attributable to a
single tier of the certificate.

{\OneAndAHalfSpacedXI
\begin{table}[ht]
\centering
\caption{Per-Tier Association Between Violation Rate and Human Consensus ($n=100$)}
\label{tab:pertier_correlation}

{\footnotesize
\renewcommand{\arraystretch}{1.2}
\setlength{\tabcolsep}{6pt}

\begin{tabular}{
>{\raggedright\arraybackslash}p{0.14\textwidth}
>{\raggedright\arraybackslash}p{0.16\textwidth}
>{\centering\arraybackslash}p{0.12\textwidth}
>{\centering\arraybackslash}p{0.12\textwidth}
>{\centering\arraybackslash}p{0.12\textwidth}
>{\centering\arraybackslash}p{0.12\textwidth}
}
\hline
\multicolumn{1}{c}{\textbf{Model Family}} &
\multicolumn{1}{c}{\textbf{Model}} &
\multicolumn{1}{c}{\textbf{T0}} &
\multicolumn{1}{c}{\textbf{T1}} &
\multicolumn{1}{c}{\textbf{T2}} &
\multicolumn{1}{c}{\textbf{T3}} \\
\hline
\multirow{2}{*}{Claude}
& \haiku  & $0.331^{***}$ & $0.477^{***}$ & $0.459^{***}$ & $0.290^{**}$ \\
& \sonnet & $0.077$        & $0.455^{***}$ & $0.368^{***}$ & $0.343^{***}$ \\
\hline
\multirow{3}{*}{GPT}
& \nano & $0.293^{**}$  & $0.294^{**}$  & $0.310^{**}$  & $0.093$ \\
& \mini & $0.368^{***}$ & $0.476^{***}$ & $0.376^{***}$ & $0.380^{***}$ \\
& \sol  & $0.095$       & $0.304^{**}$  & $0.306^{**}$  & $0.305^{**}$ \\
\hline
\multirow{2}{*}{Llama}
& \llamafast & $0.070$      & $0.293^{**}$  & $0.063$       & $0.030$ \\
& \llama     & $0.243^{*}$  & $0.494^{***}$ & $0.425^{***}$ & $0.214^{*}$ \\
\hline
\end{tabular}

\parbox{0.78\textwidth}{
\vspace{3pt}
\textit{Note.} Entries are absolute Spearman correlations, $|\rho|$, between each tier's violation rate and human consensus across the 100 prompts. Stars indicate the significance of the corresponding two-sided Spearman test: $^{*}p<0.05$, $^{**}p<0.01$, and $^{***}p<0.001$. Higher violation rates are associated with lower consensus. 
}
}
\end{table}
}

\subsection{Incremental Contribution of Certificate Tiers}\label{app:incremental}

The per-tier correlations consider each tier separately. We additionally
examine whether the distinctions introduced by successive tiers explain
variation in human consensus beyond the coarser certificate classification
that precedes them. Specifically, we compare T0 alone with the classification
obtained after incorporating T1, the T0--T1 classification with the refinement
introduced by T2, and the T0--T2 classification with the refinement introduced
by T3. Table~\ref{tab:tier_incremental} reports the resulting increases in
$R^2$ and the corresponding tests of the added distinction.

{\OneAndAHalfSpacedXI
\begin{table}[ht]
\centering
\caption{Incremental Contribution of Successive Certificate Tiers to Human Consensus}
\label{tab:tier_incremental}

{\footnotesize
\renewcommand{\arraystretch}{1.2}
\setlength{\tabcolsep}{4pt}

\begin{tabular}{
>{\raggedright\arraybackslash}p{0.13\textwidth}
>{\raggedright\arraybackslash}p{0.14\textwidth}
>{\centering\arraybackslash}p{0.09\textwidth}
>{\centering\arraybackslash}p{0.09\textwidth}
>{\centering\arraybackslash}p{0.09\textwidth}
>{\centering\arraybackslash}p{0.075\textwidth}
>{\centering\arraybackslash}p{0.09\textwidth}
>{\centering\arraybackslash}p{0.075\textwidth}
}
\hline
\noalign{\vskip 2pt}
\textbf{Model Family}
&
\textbf{Model}
&
\multicolumn{2}{c}{\textbf{T1 beyond T0}}
&
\multicolumn{2}{c}{\textbf{T2 beyond T0--T1}}
&
\multicolumn{2}{c}{\textbf{T3 beyond T0--T2}}
\\
\cmidrule(lr){3-4}
\cmidrule(lr){5-6}
\cmidrule(lr){7-8}
&
&
$\Delta R^2$
&
$p$
&
$\Delta R^2$
&
$p$
&
$\Delta R^2$
&
$p$
\\
\hline
\multirow{2}{*}{Claude}
& \haiku
& 0.136 & $<0.001^{***}$
& 0.049 & $0.018^{*}$
& 0.006 & $0.339$
\\
& \sonnet
& 0.205 & $<0.001^{***}$
& 0.022 & $0.115$
& 0.029 & $0.050$
\\
\hline
\multirow{3}{*}{GPT}
& \nano
& 0.065 & $0.013^{*}$
& 0.029 & $0.157$
& 0.015 & $0.205$
\\
& \mini
& 0.066 & $0.011^{*}$
& 0.046 & $0.025^{*}$
& 0.016 & $0.242$
\\
& \sol
& 0.091 & $<0.001^{***}$
& 0.046 & $0.029^{*}$
& 0.020 & $0.143$
\\
\hline
\multirow{2}{*}{Llama}
& \llamafast
& 0.012 & $0.302$
& 0.022 & $0.434$
& 0.022 & $0.927$
\\
& \llama
& 0.172 & $<0.001^{***}$
& 0.104 & $0.001^{**}$
& 0.020 & $0.128$
\\
\hline
\end{tabular}

\parbox{0.96\textwidth}{
\vspace{3pt}
\textit{Note.} $\Delta R^2$ reports the increase in explained variation from
adding the more refined cumulative certificate to the model containing the
immediately preceding certificate. Reported $p$-values are based on HC3
heteroskedasticity-robust tests of the added certificate coefficient.
$^{*}p<0.05$, $^{**}p<0.01$, and $^{***}p<0.001$.
}
}
\end{table}}

The largest and most consistent incremental contribution comes from T1, which
significantly improves explanatory power for six of the seven models. T2
provides additional explanatory power for several models, although the
evidence is less uniform. T3 produces smaller incremental gains and does not
yield a statistically significant contribution for any individual model under
this specification. This diminishing pattern should be interpreted in light
of the hierarchical structure of the certificate: each successive tier can
alter the classification only of recommendations that have passed all
preceding tiers.

\subsection{Alternative Explanation: Decision Difficulty}
\label{app:difficulty}

As an additional test of the warrant--consensus relationship, we examined
whether it could be explained by decision difficulty or related
characteristics of the underlying decision. Following the procedure described
in Section~\ref{sec:nomological_design}, we use log median time to first
response as a behavioral measure and estimate joint models containing warrant
and response time. Table~\ref{tab:difficulty_alternative} reports the
standardized coefficients and the incremental explanatory contribution of
each variable.

{\OneAndAHalfSpacedXI
\begin{table}[ht]
\centering
\caption{Warrant--Consensus Relationship Controlling for Decision Difficulty}
\label{tab:difficulty_alternative}

{\footnotesize
\renewcommand{\arraystretch}{1.15}
\setlength{\tabcolsep}{5pt}

\begin{tabular}{
>{\raggedright\arraybackslash}p{0.16\textwidth}
>{\raggedright\arraybackslash}p{0.14\textwidth}
>{\centering\arraybackslash}p{0.12\textwidth}
>{\centering\arraybackslash}p{0.09\textwidth}
>{\centering\arraybackslash}p{0.11\textwidth}
>{\centering\arraybackslash}p{0.10\textwidth}
>{\centering\arraybackslash}p{0.10\textwidth}
}
\hline
\noalign{\vskip 2pt}
\textbf{Model Family}
&
\textbf{Model}
&
\shortstack{\textbf{Warrant-Only}\\\textbf{Model}}
&
\multicolumn{2}{c}{\textbf{Joint Model}}
&
\multicolumn{2}{c}{\textbf{Incremental \(R^2\)}}
\\
\cmidrule(lr){3-3}
\cmidrule(lr){4-5}
\cmidrule(lr){6-7}
&
&
\(\beta_W^{\mathrm{only}}\)
&
\(\beta_W^{\mathrm{joint}}\)
&
\(\beta_{\mathrm{RT}}^{\mathrm{joint}}\)
&
\(\Delta R^2_W\)
&
\(\Delta R^2_{\mathrm{RT}}\)
\\
\hline

\multicolumn{7}{l}{\textit{Panel A: Median Time to First Response}} \\
\multirow{2}{*}{Claude}
& \haiku
& \(0.542^{***}\)
& \(0.521^{***}\)
& \(-0.067\)
& 0.247
& 0.005
\\
& \sonnet
& \(0.480^{***}\)
& \(0.455^{***}\)
& \(-0.146\)
& 0.201
& 0.021
\\
\hline
\multirow{3}{*}{GPT}
& \nano
& \(0.358^{***}\)
& \(0.329^{***}\)
& \(-0.169\)
& 0.106
& 0.028
\\
& \mini
& \(0.524^{***}\)
& \(0.500^{***}\)
& \(-0.136\)
& 0.242
& 0.018
\\
& \sol
& \(0.389^{***}\)
& \(0.365^{***}\)
& \(-0.174\)
& 0.131
& 0.030
\\
\hline
\multirow{2}{*}{Llama}
& \llamafast
& \(0.186^{*}\)
& \(0.163^{*}\)
& \(-0.206^{*}\)
& 0.026
& 0.042
\\
& \llama
& \(0.596^{***}\)
& \(0.575^{***}\)
& \(-0.085\)
& 0.312
& 0.007
\\

\hline
\multicolumn{7}{l}{\textit{Panel B: Median Total Response Time}} \\
\multirow{2}{*}{Claude}
& \haiku
& \(0.542^{***}\)
& \(0.522^{***}\)
& \(-0.073\)
& 0.252
& 0.005
\\
& \sonnet
& \(0.480^{***}\)
& \(0.457^{***}\)
& \(-0.143\)
& 0.204
& 0.020
\\
\hline
\multirow{3}{*}{GPT}
& \nano
& \(0.358^{***}\)
& \(0.334^{***}\)
& \(-0.167\)
& 0.109
& 0.027
\\
& \mini
& \(0.524^{***}\)
& \(0.502^{***}\)
& \(-0.138\)
& 0.246
& 0.019
\\
& \sol
& \(0.389^{***}\)
& \(0.368^{***}\)
& \(-0.171\)
& 0.134
& 0.029
\\
\hline
\multirow{2}{*}{Llama}
& \llamafast
& \(0.186^{*}\)
& \(0.163^{*}\)
& \(-0.197\)
& 0.026
& 0.039
\\
& \llama
& \(0.596^{***}\)
& \(0.576^{***}\)
& \(-0.088\)
& 0.316
& 0.007
\\
\hline
\end{tabular}

\parbox{0.96\textwidth}{
\vspace{3pt}
\textit{Note.} Human consensus, warrant, and response time are standardized
to mean 0 and standard deviation 1. Both response-time measures are
log-transformed. The Warrant-Only Model reports the warrant coefficient from
a specification containing only warrant. The Joint Model includes both
warrant and response time. \(\beta_W^{\mathrm{only}}\) is the warrant coefficient from the warrant-only
model. \(\beta_W^{\mathrm{joint}}\) and
\(\beta_{\mathrm{RT}}^{\mathrm{joint}}\) are the warrant and response-time
coefficients from the joint model. \(\Delta R^2_W\) is the additional variation in
human consensus explained by warrant beyond response time, whereas
\(\Delta R^2_{\mathrm{RT}}\) is the additional variation explained by response
time beyond warrant. Reported significance levels are based on HC3
heteroskedasticity-robust tests. Each specification uses \(n=100\).
\(^{*}p<0.05\), \(^{**}p<0.01\), and \(^{***}p<0.001\).
}
}
\end{table}}

Across all seven models, the warrant coefficient remains positive when
decision difficulty is included, and it is statistically significant in every
case. Warrant also explains additional variation in human consensus beyond
difficulty for every model. In contrast, difficulty is statistically
significant for only \llamafast\ and generally contributes comparatively little
incremental explanatory power. These results indicate that the observed
warrant--consensus relationship is not readily explained by differences in
the difficulty of the underlying decisions.

\section{Additional Results for Discriminant Validity}
\label{app:discriminant}

As an additional test of discriminant validity, we examine whether warrant explains variation in human consensus beyond that captured by verbalized confidence. For each model, we estimate a baseline regression relating ranked human consensus to ranked verbalized confidence and an extended regression that additionally includes ranked warrant category. This analysis complements the correlation-based evidence in Section~\ref{sec:discriminant_results} by  asking whether warrant contains information relevant to human consensus that is not already captured by confidence. Table~\ref{tab:incremental} reports the baseline and extended $R^2$, the resulting increase in explained variation from adding warrant, and the HC3-robust test of the warrant coefficient for each model.

{\OneAndAHalfSpacedXI
\begin{table}[ht]
\centering
\captionof{table}{Incremental Explanatory Value of Warrant for Human Consensus Beyond Verbalized Confidence}
\label{tab:incremental}

{\footnotesize
\renewcommand{\arraystretch}{1.2}
\setlength{\tabcolsep}{6pt}

\begin{tabular}{
>{\raggedright\arraybackslash}p{0.14\textwidth}
>{\raggedright\arraybackslash}p{0.14\textwidth}
>{\centering\arraybackslash}p{0.12\textwidth}
>{\centering\arraybackslash}p{0.14\textwidth}
>{\centering\arraybackslash}p{0.10\textwidth}
>{\centering\arraybackslash}p{0.10\textwidth}
}
\hline
\textbf{Model Family} & \textbf{Model} & \textbf{Baseline $R^2$} & \textbf{Extended $R^2$} & \textbf{$\Delta R^2$} & \textbf{$p$} \\
\hline
\multirow{2}{*}{Claude}
& \haiku  & 0.244 & 0.342 & 0.098 & $<0.001^{***}$ \\
& \sonnet & 0.310 & 0.353 & 0.044 & $0.017^{*}$ \\
\hline
\multirow{3}{*}{GPT}
& \nano & 0.084 & 0.181 & 0.097 & $0.001^{**}$ \\
& \mini & 0.149 & 0.297 & 0.148 & $<0.001^{***}$ \\
& \sol  & 0.202 & 0.234 & 0.032 & $0.063$ \\
\hline
\multirow{2}{*}{Llama}
& \llamafast & 0.012 & 0.027 & 0.015 & $0.210$ \\
& \llama     & 0.158 & 0.351 & 0.194 & $<0.001^{***}$ \\
\hline
\end{tabular}

\parbox{0.87\textwidth}{
\vspace{3pt}
\textit{Note.} Human consensus is the dependent variable. The dependent variable and predictors are rank-transformed. Baseline models include verbalized confidence only; extended models additionally include warrant category. $\Delta R^2$ denotes the increase in explained variation in human consensus from adding warrant. Reported $p$-values correspond to HC3 heteroscedasticity-robust tests of the warrant coefficient in the extended model. $^{*}p<0.05$, $^{**}p<0.01$, and $^{***}p<0.001$.
}
}
\end{table}}

\section{Additional Robustness Results}
\label{app:robustness}

\subsection{Alternative Certificate Operationalizations}
\label{app:alternative_cert}

Our primary analyses represent the certificate using the four-level warrant classification. To assess whether the warrant--consensus relationship depends on this particular discretization, we repeat the analysis using two continuous operationalizations of certificate strength: a lexicographic score that preserves the tier hierarchy while incorporating within-tier violation rates, and a pooled score based on violations across all certificate queries. Both measures are reverse-coded so that higher values indicate stronger warrant.

{\OneAndAHalfSpacedXI
\begin{table}[ht]
\centering
\captionof{table}{Robustness of the Warrant--Consensus Association Under Alternative Certificate Operationalizations ($n=100$)}
\label{tab:robustness_correlation}

{\footnotesize
\renewcommand{\arraystretch}{1.2}
\setlength{\tabcolsep}{6pt}

\begin{tabular}{
>{\raggedright\arraybackslash}p{0.14\textwidth}
>{\raggedright\arraybackslash}p{0.16\textwidth}
>{\centering\arraybackslash}p{0.18\textwidth}
>{\centering\arraybackslash}p{0.10\textwidth}
>{\centering\arraybackslash}p{0.14\textwidth}
>{\centering\arraybackslash}p{0.10\textwidth}
}
\hline
\textbf{Model Family} & \textbf{Model} & \textbf{$\rho$ (Lexicographic)} & \textbf{$p$} & \textbf{$\rho$ (Pooled)} & \textbf{$p$} \\
\hline
\multirow{2}{*}{Claude}
& \haiku & 0.568 & $<0.001^{***}$ & 0.496 & $<0.001^{***}$ \\
& \sonnet & 0.510 & $<0.001^{***}$ & 0.505 & $<0.001^{***}$ \\
\hline
\multirow{3}{*}{GPT}
& \nano & 0.388 & $<0.001^{***}$ & 0.338 & $<0.001^{***}$ \\
& \mini & 0.544 & $<0.001^{***}$ & 0.501 & $<0.001^{***}$ \\
& \sol & 0.391 & $<0.001^{***}$ & 0.390 & $<0.001^{***}$ \\
\hline
\multirow{2}{*}{Llama}
& \llamafast & 0.289 & $0.004^{**}$ & 0.199 & $0.047^{*}$ \\
& \llama & 0.620 & $<0.001^{***}$ & 0.540 & $<0.001^{***}$ \\
\hline
\end{tabular}

\parbox{0.95\textwidth}{
\vspace{3pt}
\textit{Note.} Spearman's $\rho$ measures the association between each alternative certificate operationalization and human consensus across the 100 prompts. Both alternative measures are reverse-coded so that higher values indicate stronger warrant. Higher values are therefore expected to be associated with greater human consensus. $^{*}p<0.05$, $^{**}p<0.01$, and $^{***}p<0.001$.
}
}
\end{table}}

As shown in Table~\ref{tab:robustness_correlation}, the association with human consensus is preserved under both alternative operationalizations. All seven models exhibit significant positive associations under the lexicographic measure, including \llamafast, for which the coarse four-category warrant measure produces a weaker relationship. The pooled measure yields the same qualitative pattern, with significant positive associations for all seven models. These results indicate that the warrant--consensus relationship is not an artifact of the particular categorical representation used in the primary analysis.

\subsection{Alternative Transformation-Generation Procedure}
\label{app:alternative_transformations}

To assess whether the main validation results depend on the particular procedure used to generate certificate transformations, we repeated the analysis using the alternative transformation-generation procedure described in Appendix~\ref{app:alternative_procedure}. 

Table~\ref{tab:alt_transform_robustness} reports the association between warrant and human consensus under this alternative procedure. The relationship remains positive and statistically significant for all six models examined, indicating that the  result is not specific to the original transformation-generation procedure.

\begin{table}[ht]
\centering
{\OneAndAHalfSpacedXI
\caption{Robustness of the Warrant--Consensus Relationship Under Alternative Transformations}
\label{tab:alt_transform_robustness}

{\footnotesize
\renewcommand{\arraystretch}{1.2}
\setlength{\tabcolsep}{6pt}

\begin{tabular}{
>{\raggedright\arraybackslash}p{0.14\textwidth}
>{\raggedright\arraybackslash}p{0.19\textwidth}
>{\centering\arraybackslash}p{0.13\textwidth}
>{\centering\arraybackslash}p{0.13\textwidth}
}
\hline
\multicolumn{1}{c}{\textbf{Model Family}} &
\multicolumn{1}{c}{\textbf{Model}} &
\multicolumn{1}{c}{\textbf{Spearman $\rho$}} &
\multicolumn{1}{c}{\textbf{$p$}} \\
\hline
\multirow{2}{*}{Claude}
& \haiku  & 0.384 & $<0.001^{***}$ \\
& \sonnet & 0.404 & $<0.001^{***}$ \\
\hline
GPT
& \nano & 0.278 & $0.005^{**}$ \\
& \mini & 0.443 & $<0.001^{***}$ \\
& \sol  & 0.283 & $0.004^{**}$ \\
\hline
Llama
& \llamafour & 0.293 & $0.003^{**}$ \\
\hline
\end{tabular}

\parbox{0.68\textwidth}{
\vspace{3pt}
\textit{Note.} Spearman's $\rho$ measures the association between warrant category and human consensus across the 100 prompts using the alternative transformation-generation procedure. $^{*}p<0.05$, $^{**}p<0.01$, and $^{***}p<0.001$.
}
}
}
\end{table}

\end{APPENDICES}
\end{document}